\documentclass[a4paper,11pt]{article}
\usepackage{amssymb,amsfonts,amsthm,amsmath}  %% <- after lineno
\usepackage[bookmarks=false]{hyperref}
\usepackage[bbgreekl]{mathbbol}
\usepackage{tikz}
\usepackage[backend=biber, style=numeric, citestyle=ieee]{biblatex}
\usetikzlibrary{arrows.meta, decorations.text}
\usetikzlibrary{decorations.pathmorphing,patterns}
\usetikzlibrary{calc}
\usepackage{pgfplots}
\usepgfplotslibrary{fillbetween}
\usepackage{stmaryrd}
\usepackage{authblk}
\usepackage{thmtools, thm-restate}
\usepackage{float}
\usepackage{mathtools}

\usepackage{stackengine}
\usepackage{algorithm}
\usepackage{algpseudocode} 
\usepackage{subfigure}
\usepackage{graphicx}
\usepackage{textcomp}
\usepackage[switch, modulo]{lineno}
\usepackage{array, multirow, bigdelim, makecell, booktabs} 
\hypersetup{
    unicode=true,          % non-Latin characters in Acrobat’s bookmarks
    pdftoolbar=true,        % show Acrobat’s toolbar?
    pdfmenubar=true,        % show Acrobat’s menu?
    pdffitwindow=false,     % window fit to page when opened
    pdfstartview={FitH},    % fits the width of the page to the window
    pdftitle={My title},    % title
    pdfauthor={Author},     % author
    pdfsubject={Subject},   % subject of the document
    pdfcreator={Creator},   % creator of the document
    pdfproducer={Producer}, % producer of the document
    pdfkeywords={keyword1, key2, key3}, % list of keywords
    pdfnewwindow=true,      % links in new PDF window
    colorlinks=true,       % false: boxed links; true: colored links
    linkcolor=red!85!black,          % color of internal links (change box color with linkbordercolor)
    citecolor=red!85!black,        % color of links to bibliography
    filecolor=cyan,         % color of file links
    urlcolor=red!85!black        % color of external links
}

\DeclareSymbolFontAlphabet{\mathbbl}{bbold}
\allowdisplaybreaks[1]
\newcommand{\NJ}{\textcolor{black}}

\title{\NJ{Tunable correlation retention}: A statistical method for generating synthetic data}
\author[1,*]{Nicklas J\"averg\aa rd}
\author{Adrian Muntean, Rainey Lyons}
\affil[1]{Department of Mathematics and Computer Science, Karlstad University, Sweden}
\author{Jonas Forsman}
\affil[2]{CGI, Data Advantage, Karlstad, Sweden}
\affil[*]{Correspondence to nicklas.javergard@kau.se}
\date{}

\pgfplotsset{compat=1.18}
\begin{document}

\maketitle
\begin{abstract}
\noindent

We propose a method to generate statistically representative synthetic data from a given dataset. 
The main goal of our method is for the created data set to mimic the inter--feature correlations present in the original data, while also offering a tunable parameter to influence the privacy level. 
In particular, our method constructs a statistical map by using the empirical conditional distributions between the features of the original dataset. Part of the tunability is achieved by limiting the depths of conditional distributions that are being used.
%The main goal is to be able to maintain in the synthetic dataset the correlations of the features present in the original one, while offering the possibility of a tunable privacy level.
We describe in detail our algorithms used both in the construction of a statistical map and how to use this map to generate synthetic observations. 
%For the construction of the statistical map we use empirical conditional distributions between the features of the original dataset.
\NJ{This approach is tested in three different ways: with a hand calculated example; a manufactured dataset; and a real world energy-related dataset of consumption/production of households in Madeira Island. 
We evaluate the method by comparing the datasets using the Pearson correlation matrix with different levels of resolution and depths of correlation. These two considerations are being viewed as tunable parameters influencing the resulting datasets fidelity and privacy. }

The  proposed methodology is general in the sense that it does not rely on the used test dataset. We expect it to be applicable in a much broader context than indicated here.

\end{abstract}

\section{Introduction}

Computational science and engineering are constantly facing new data-related challenges, some of which involve contradictory requests. 
For instance, in the presence of complex projects such as the creation of digital twins of manufacturing processes, the development of realistic smart cities, or the reliable prediction of trends in the evolution of energy consumption or financial markets in the presence of uncertainties, researchers need access to large high-quality datasets for data-driven modeling; see, e.g., the recent review article \cite{Vishnu}. 
On the other hand, data owners are often hesitant to share detailed information due to various concerns. 
One common example of such concerns is with respect to privacy, particularly regarding sensitive data such as medical records, survey responses, or household-level electrical consumption. 
The latter of which is of increasing importance in the context of system monitoring and load prediction. 
Additionally, companies storing extensive datasets are concerned about disclosing information that could adversely affect their competitiveness.
Such developments make the generation of synthetic data with tuneable fidelity and privacy critical. 
In this work, we propose a simple method of synthetic data generation that allows a tunable quality of statistical information which sets the stage for a controllable level of privacy. 

\NJ{In the literature, there are two broad approaches to generating synthetic data. One utilizes different neural network structures such as, generative adversarial networks (GANs), variational autoencoders (VAEs), convolutional neural networks (CNNs), long short-term memory (LSTM), or recurrent neural networks (RNNs) to name a few. % If we take a closer look for instance at GANs, which does a good job in generating synthetic images, they have room to improve when it comes to tabular data according to \cite{GANSurvey2022}.
Most machine learning approaches do not allow introspection of the way decisions are being made as the synthetic data is being generated. This is not true for classification and regression trees originally proposed in \cite{Reiter2005} and applied to generate synthetic data in \cite{pater2023}. 
On the contrary, classical methods like oversampling, rotation, scaling, interpolation, and Bayesian networks all allow the process of generation to be inspected. 
Of these classical methods, Bayesian networks comes closest to what we are doing in this work. They construct probabilistic models that represent variables and their conditional probabilities as directed acyclic graphs, for more information see \cite{young2009using} and \cite{gogoshin2021synthetic}. }

\NJ{The method to follow does not propose any preexisting structure. 
Instead, we compute all the conditional distributions up to some prescribed depth and randomly select paths as the synthetic data is generated. 
We discuss our methodology in section \ref{method_overview} and apply it on 3 distinct datasets which are referred to as ``original" and are denoted generically by $\mathcal{O}$.}

This paper is structured in the following way.
In Section \ref{method_overview}, we present by means of an example and a formulation in simple mathematical terms how our method works and what it must deliver, while the corresponding algorithms and a few implementation details are the subject of Section \ref{Implementation}. 
Using a particular large dataset collected from the energy sector, introduced in Section \ref{acquisition}, we focus our attention in Section \ref{current_results} on qualitative and quantitative results obtained when comparing our synthetically generated datasets with the original dataset. 
Finally, in Section \ref{final_discussion} we discuss the obtained results and anticipate as well further potential developments of our method for synthetic data generation.

\section{Description of the method}\label{method_overview}

\subsection{Example by hand calculation}
Before describing the method in general, we explain the idea behind our algorithm by means of an example where all computations can be done by hand. We first describe the calculation for the conditional probabilities between combinations of features and their discretizations. Using the computed probabilities, we then generate synthetic data. \NJ{In the next subsection we formulate the method in generality.} 
\subsubsection{Estimation of conditional probabilities} \label{sec:HandCalc_1}

\NJ{Let $\mathcal{O}$ denote our original dataset of size $S=6$. 
We view each observation as a vector in $\mathbb{R}^3$, i.e., }
\begin{equation}
    \mathcal{O} = \{(f_1^{(s)}, f_2^{(s)}, f_3^{(s)}) : s = 1, \dots, 6 \}  \subset \mathbb{R}^{S\times 3},
\end{equation}
\NJ{where each $f_j^{(s)}$ is an independent realization of the random variables $F_1, F_2$ or $F_3$, respectively, with the distributions }
\begin{equation}
    F_1 \sim \mathcal{U}([0,2]), \qquad F_2\sim \mathcal{U}([0,1]), \qquad F_3\sim\mathcal{U}([0,0.5]),
\end{equation}
where $\mathcal{U}([a, b])$ for $a<b$ is the uniform distribution on $[a, b]$. \\
\NJ{For this particular example, say the realization is given by}
\begin{equation*}
    \mathcal{O} = 
    \begin{array}{|c|c|c|}
        f_1& f_2& f_3 \\
         1.75& 0.23& 0.03 \\
         0.75& 0.05& 0.26 \\
         0.54& 0.82& 0.40 \\
         0.84& 0.04& 0.36 \\
         0.80& 0.76& 0.14 \\
         0.91& 0.68& 0.30 \\
    \end{array},
\end{equation*}
\NJ{where $f_i:=\{f_i^{(s)}\}_{s=1}^6$. We focus on each $f_i$ separately and define the empirical minimum and maximum by}

\begin{equation}
    m_i := \min_{1\leq s\leq 6} f_i^{(s)}, \qquad M_i := \max_{1\leq s \leq 6} f_i^{(s)}.
    \label{eq:minMax}
\end{equation}

\NJ{For a given $N$ (in this example take $N=4$), we partition the interval $[m_i, M_i]$ into $N$ disjoint subintervals of equal length}

\begin{equation}
    \Delta_i := \frac{M_i-m_i}{N}.
    \label{eq:Delta}
\end{equation}

\NJ{For $n= 1, \dots, N$, we denote each subinterval by $A_i^n$, i.e.,}

\begin{equation}
    A_i^n = \begin{cases}
        [m_i + (n-1)\Delta_i, m_i + n\Delta_i],&n=1, \dots, N-1, \\
        [m_i + (N-1)\Delta_i, M_i], & n=N, 
    \end{cases}
    \label{eq:intervals}
\end{equation}
or more explicitly,
\begin{align*}
    &A_1^1=[0.54, 0.85), &&A_1^2=[0.85, 1.14), &&&A_1^3=[1.14, 1.45), &&&&A_1^4=[1.45, 1.75], \\
    &A_2^1=[0.04, 0.23), &&A_2^2=[0.23, 0.43), &&&A_2^3=[0.43, 0.63), &&&&A_2^4=[0.63, 0.82], \\
    &A_3^1=[0.03, 0.12), &&A_3^2=[0.12, 0.21), &&&A_3^3=[0.21, 0.31), &&&&A_3^4=[0.31, 0.40]. 
\end{align*}
\NJ{The estimation of probability distribution of each $f_i$ is done by means of the relative frequency along its four subintervals.
This procedure applied to $f_1$ results in Figure \ref{fig:HandCalc_Order1}.
The heights of each pillar represent the probability of a value taken from $\mathcal{O}$ to be in a interval $A_1^n$, $n\in\{1, 2, 3, 4\}$.}

\begin{figure}
    \centering
    \begin{tikzpicture}
        \draw[thick, black, ->] (0,0) -- (4.5,0);
        \draw[thick, black, ->] (0,0) -- (0,3);
        \filldraw[color=black!100, fill=blue!40, thick] (0.0,0.0) rectangle (1,2);
        \filldraw[color=black!100, fill=blue!40, thick] (1,0.0) rectangle (2,1);
        \filldraw[color=black!100, fill=blue!40, thick] (3,0.0) rectangle (4,1);
        \draw[black, thin] (0,1) -- (-0.1, 1);
        \draw[black, thin] (0,2) -- (-0.1, 2);
        \node at (-0.5, 1) {$1/6$};
        \node at (-0.5, 2) {$4/6$};
        \node at (0.5,-0.3) {$A_1^1$};
        \node at (1.5,-0.3) {$A_1^2$};
        \node at (2.5,-0.3) {$A_1^3$};
        \node at (3.5,-0.3) {$A_1^4$};
    \end{tikzpicture}
    \caption{Estimation of the empirical distribution of $f_1$.}
    \label{fig:HandCalc_Order1}
\end{figure}
\NJ{\noindent Given that a value in $A_1^1$ was observed, it forces the possible values in $f_2$ to be, $0.05, 0.82, 0.04,$ and $0.75$ since they correspond to the rows of $f_1$ in $A_1^1$. By repeating the procedure as described above to this subset of $f_2$ leads to Figure \ref{fig:HandCalc_Order2}. Theoretically, this procedure can be repeated until all columns (realization of the corresponding random variable) are used. For this example, we do it three times.  }

\begin{figure}
    \centering
    \begin{tikzpicture}
        \draw[thick, black, ->] (0,0) -- (4.5,0);
        \draw[thick, black, ->] (0,0) -- (0,3);
        \filldraw[color=black!100, fill=blue!40, thick] (0.0,0.0) rectangle (1,2);
        \filldraw[color=black!100, fill=blue!40, thick] (3,0.0) rectangle (4,2);
        \draw[black, thin] (0,2) -- (-0.1, 2);
        \node at (-0.5, 2) {$1/2$};
        \node at (0.5,-0.3) {$A_2^1$};
        \node at (1.5,-0.3) {$A_2^2$};
        \node at (2.5,-0.3) {$A_2^3$};
        \node at (3.5,-0.3) {$A_2^4$};
    \end{tikzpicture}
    \caption{Estimation of the empirical conditional distribution of $f_2$ given that the observed feature one is in the subinterval $A_1^1$, i.e., $f_1 \in A_1^1$.}
    \label{fig:HandCalc_Order2}
\end{figure}
\NJ{\noindent The last step, given a value in $A_1^1$ of $f_1$ and a value in $A_2^4$ of $f_2$, it forces the possible values of $f_3$ to be $0.40$ and $0.14$. Thus the result is illustrated in Figure \ref{fig:HandCalc_Order3}. }

\begin{figure}
    \centering
    \begin{tikzpicture}
        \draw[thick, black, ->] (0,0) -- (4.5,0);
        \draw[thick, black, ->] (0,0) -- (0,3);
        \filldraw[color=black!100, fill=blue!40, thick] (1,0.0) rectangle (2,2);
        \filldraw[color=black!100, fill=blue!40, thick] (3,0.0) rectangle (4,2);
        \draw[black, thin] (0,2) -- (-0.1, 2);
        \node at (-0.5, 2) {$1/2$};
        \node at (0.5,-0.3) {$A_3^1$};
        \node at (1.5,-0.3) {$A_3^2$};
        \node at (2.5,-0.3) {$A_3^3$};
        \node at (3.5,-0.3) {$A_3^4$};
    \end{tikzpicture}
    \caption{Estimation of the empirical conditional distribution of $f_3$ given that $f_1 \in A_1^1$ and $f_2 \in A_2^4$.}
    \label{fig:HandCalc_Order3}
\end{figure}
\noindent By applying this approach to all columns and bins of a dataset, the resulting empirical distributions can be used to generate synthetic data by sampling. 

\subsubsection{Generating synthetic data}
\NJ{The standing assumption is that the estimation of the conditional probabilities in section \ref{sec:HandCalc_1} is done for all features and combinations thereof. Generation of one synthetic observation is as follows: Select a feature randomly. Following the previous section, say the chosen feature is $f_1$. Then, randomly select one of the subintervals $\{A_1^n\}_{n=1}^4$  according to the empirical probability distribution shown in Figure \ref{fig:HandCalc_Order1}. 
Given that this interval is $A_1^1$, draw a random number, $x$,  from $\mathcal{U}(A_1^1)$. Since $x\in A_1^1$, we select a value for feature $f_2$ from one of the subintervals $\{A_2^n\}_{n=1}^4$ which is chosen randomly according to the empirical \textit{conditional} probability distribution shown in Figure \ref{fig:HandCalc_Order2}. Given that this interval is $A_2^4$, draw a random number, $y$, from $\mathcal{U}(A_2^4)$. Since $x\in A_1^1$ and $y\in A_2^4$, select one of $A_3^2$ or $A_3^4$, randomly according to the empirical conditional probability distribution shown in Figure \ref{fig:HandCalc_Order3}. Given that this interval is $A_3^2$, draw a random number, $z$, from $\mathcal{U}(A_3^3)$. 
The set $\{x, y, z\}$ is then a synthetic observation, i.e. a new row in a synthetic dataset representing the original.
We can then repeat this process until the desired number of observations is created. }

\subsection{General formulation} \label{sec:GenFormulation}

\NJ{We consider an original dataset $\mathcal{O}$ with $S$ independent realizations of $N_f$ real-valued variables such that}
\begin{equation}
    \mathcal{O} = \{(f_1^{(s)}, f_2^{(s)}, \dots, f_{N_f}^{(s)})\in\mathbb{R}^{N_f} : s = 1, \dots, S \} \in \mathbb{R}^{S\times N_f},
\end{equation}
\NJ{where $f_i^{(s)}$ represents the observed value of the random variable $F_i$ for $i\in\{1, \dots, N_f\}$. The set of realizations is denoted by }
\begin{equation}
    f_i := \{f_i^{(s)}\}_{s=1}^M.
\end{equation}
\NJ{For each $f_i$ we apply equation \eqref{eq:minMax} and \eqref{eq:Delta} in order to arrive at the subintervals for each variable as shown in \eqref{eq:intervals}.}
To streamline the presentation, we introduce the following convention for indices: $i, j, k \in\{1, \dots, N_f\}$ and $n, m, \ell\in \{0, \dots, N\}$. 
\NJ{Each random variable $F_i$, with an unknown distribution $P_i$, is assumed to admit a continuous density function $\rho_i(f_i)$ with respect to the Lebesgue measure.\footnote{\NJ{In the current implementation of our method we do not deal with categorical or discrete data. In principal both are manageable even through the categorical data would have to be encoded numerically.}}}
\NJ{The probability of a value in $A^n_i$ would then be given by }
\NJ{
\begin{equation}
    p_{i, n} = P(F_i\in A^n_i) = \int_{A^n_i} \rho_i(f_i)df_i
    \label{eq:probofset}
\end{equation}
}
%\begin{equation}
%    p_{i, n} =P(A^n_i) = \int_{A^n_i}\rho_i(x) dx.
%    \label{eq:probofset}
%\end{equation}
We can estimate $p_{i,n}$ with the relative frequency of points in the interval $A^n_i$ via
\begin{equation}
    \hat{p}_{i,n} = \frac{1}{M}\sum_{j=1}^{M}\chi_{A_i^n}(f_i^j),
    \label{eq:emp_prob}
\end{equation}
where $\chi_A(\cdot)$ is the indicator function of event $A$ defined by 

\begin{equation}
    \chi_A(x):=
    \begin{cases}
        1 & \text{if } x\in A, \\
        0 & \text{if } x\notin A.
    \end{cases}
\end{equation}
The probability of a set $A\subseteq \bigcup_{n=0}^N A_i^n$ is approximated by means of the following expression

\begin{equation}
    \int_A \rho_i(x) dx \approx \sum_{\{n : A_i^n\cap A\neq \emptyset\}}\hat{p}_{i,n}.
    \label{eq:measure_mu_i}
\end{equation}
Suppose that $(x, y, z) \in (f_i, f_j, f_k)$ such that $i\neq j\neq k$. We are concerned with the estimation of two types of conditional probabilities. Specifically,  we are interested in the first order conditional probability, i.e. the probability that $y\in A_j^m$ given that $x\in A_i^n$ and the second order conditional probability, i.e. the probability that $z\in A_k^\ell$ given that $x\in A_i^n$ and $y \in A_j^m$, or more concisely, in $P(F_j\in A_j^m|F_i\in A_i^n)$ and $P(F_k\in A_k^\ell|F_i \in A_i^n, F_j\in A_j^m).$ Looking at the first order conditional probabilities,
\NJ{we want to estimate the quantity}

\begin{equation*}
    P(F_j\in A_j^m|F_i \in A_i^n) = \frac{P(F_j\in A_j^m, F_i \in A_i^n)}{P(F_i \in A_i^n)} = \frac{P((F_j, F_i)\in(A_j^m\times  A_i^n))}{P(F_i \in A_i^n)},
    \label{eq:cond_theoreticalprob}
\end{equation*}
where $P(F_j\in A_j^m, F_i \in A_i^n)$ means $P(F_j\in A_j^m\text{ and }F_i \in A_i^n)$.\\
\NJ{The joint distribution}

\begin{equation}
    p_{[(j,i), (m,n)]} := P((F_j, F_i)\in(A_j^m\times  A_i^n)),
\end{equation}
\NJ{can be estimated by}

\begin{equation}
    \hat{p}_{[(j,i), (m,n)]} = \frac{1}{M} \sum_{s=1}^{M} \chi_{A_j^m\times A_i^n}(f_j^{(s)}, f_i^{(s)}).
    \label{eq:emp_jointprob}
\end{equation}
By combining \eqref{eq:emp_jointprob} and \eqref{eq:emp_prob}, an estimate of

\begin{equation}
    p_{[(j,m)|(i,n)]} = P(F_j\in A_j^m|F_i \in A_i^n)
\end{equation}
can be expressed as
\begin{equation}
    \hat{p}_{[(j,m)|(i,n)]} = \frac{\hat{p}_{[(j,i), (m.n)]}}{\hat{p}_{i,n}} = \frac{ \sum_{s=1}^{M} \chi_{A_j^m\times A_i^n}(f_j^{(s)}, f_i^{(s)})}{\sum_{s=1}^{M}\chi_{A_i^n}(f_i^{(s)})}. 
    \label{eq:emp_condprob}
\end{equation}
\NJ{The tri-variate joint probability is given by}

\begin{equation}
    p_{[(k,j,i), (\ell, m, n)]} = P((F_k,F_j,F_i)\in (A_k^\ell\times A_j^m \times A_i^n))
    \label{eq:theo_jointprob_2}
\end{equation}
and can be estimated by

\begin{equation}
    \hat{p}_{[(k,j,i), (\ell,m,n)]} = \frac{1}{M}\sum_{s=1}^M \chi_{A_k^\ell\times A_j^m \times A_i^n}(f_k^{(s)},f_j^{(s)}, f_i^{(s)}).
    \label{eq:emp_jointprob_2}
\end{equation}
Combining \eqref{eq:emp_jointprob_2} and \eqref{eq:emp_jointprob}, an estimate of

\begin{equation}
    p_{[(k,\ell)|(i,n),(j,m)]} = P(F_k \in A_k^\ell|F_j\in A_j^m,F_i\in A_i^n)
\end{equation}
can be expressed as

\begin{equation}
    \hat{p}_{[(k,\ell)|(j,m),(i,n)]} = 
    \frac{\hat{p}_{[(k,j,i), (\ell,m,n)]}}{\hat{p}_{[(j,i), (m,n)]}}=
    \frac{\sum_{s=1}^M \chi_{A_k^\ell\times A_j^m \times A_i^n}(f_k^{(s)},f_j^{(s)}, f_i^{(s)})}{ \sum_{s=1}^{M} \chi_{A_j^m\times A_i^n}(f_j^{(s)}, f_i^{(s)})}. 
    \label{eq:emp_condprob_2}
\end{equation}
Using \eqref{eq:emp_prob}, \eqref{eq:emp_jointprob}, and \eqref{eq:emp_jointprob_2}, the corresponding density functions can be approximated with histograms that are constant over each set of type $A_i^k$, $A_i^m\times A_i^n$ and $A_k^\ell\times A_j^m \times A_i^n$, respectively. Denoting the length of the interval $A_i^n$ by $|A_i^n|$, the height of the bars in each set is given by

\begin{equation}
    h_i^k := \frac{\hat{p}_{i,k}}{|A_i^k|},
\end{equation}
in the 1D case, and by

\begin{equation}
    h_{[(j,i), (m,n)]} := \frac{\hat{p}_{[(j,i), (m,n)]}}{|A_j^m||A_i^n|},
\end{equation}
in the 2D case, and by

\begin{equation}
    h_{[(k,j,i),(\ell,m,n)]} := \frac{\hat{p}_{[(k,j,i), (\ell,m,n)]}}{|A_k^\ell||A_j^m||A_i^n|}.
\end{equation}
in the 3D case.
Using $h^k_i$, $h_{[(j,i),(m,n)]}$ and $h_{[(k,j,i),(\ell,m,n)]}$, the uni-, bi- and tri-variate probability density functions can now be constructed. 
We are concluding this section with a brief description of the generation of synthetic data.

Given the uni-, bi- and tri-variate density functions approximated by histograms, the synthetic data can be generated in three steps:
\begin{enumerate}
    \item  select an interval $A_i^n$ randomly, according to its probability mass, draw a random number from $\mathcal{U}(A_i^n)$;
    \item select an interval $A_j^m$ according to it probability mass conditioned on the first interval $A_i^n$, draw a number from $\mathcal{U}(A_j^m)$;
    \item select an interval, $A_k^\ell$, according to its probability mass conditioned on both $A_i^n$ and $A_j^m$ and draw a value from $\mathcal{U}(A_k^\ell)$.
\end{enumerate}
If the dataset at hand contains more than three features, only the last step is repeated until a full row of the synthetic dataset is generated. Meaning that the two first intervals are reused to be conditions for any other feature that still miss a value in the synthetic dataset.

\section{Implementation}\label{Implementation}

Herein we elucidate the workflow of the proposed algorithms corresponding to the methodology proposed in section \ref{sec:GenFormulation}. Algorithm \ref{alg:map1} and \ref{alg:map2} show the steps for computing the estimates of the first and the second order conditional probabilities, respectively. Algorithm \ref{alg:gen1} and \ref{alg:gen2} show the steps for the generation of synthetic data assuming the steps of Algorithm \ref{alg:map1} or \ref{alg:map2} have been done. All implementations are written in \textit{Julia} \cite{bezanson2017julia}.

\begin{algorithm}[H]

\caption{ Implementation for estimating the first order conditional probabilities of an original dataset} \label{alg:map1}

\begin{algorithmic}[1]
\For{\texttt{each $f_i$}}
    \State \texttt{Discretize the range of feature $f_i$, forming intervals $A^n_i$}
    \For{each $A^n_i$}
        \State\texttt{Compute $P(F_i \in A^n_i)$}
    \EndFor
\EndFor
\For{\texttt{each $f_i$}}
    \For{\texttt{each $f_j : f_j\neq f_i$}}
        \For{\texttt{each $A^n_i\in \{A^1_i, \dots, A_i^N\}$}}
            \For{\texttt{each $A^m_j\in\{A^1_j,\dots,A^N_j\}$}}
                \State\texttt{Compute $P(F_j\in A^m_j\vert F_i\in A^n_i)$}
            \EndFor
        \EndFor
    \EndFor
\EndFor
\end{algorithmic}
\end{algorithm}

\begin{algorithm}[H]

\caption{Implementation for estimating the first and second order conditional probabilities of an original dataset} \label{alg:map2}

\begin{algorithmic}[1]
\For{\texttt{each $f_i$}}
    \State \texttt{Discretize the range of feature $f_i$, forming intervals $A^n_i$}
    \For{\texttt{each $A^n_i\in \{A^1_i, \dots, A_i^N\}$}}
        \State\texttt{Compute $P(F_i\in A^n_i)$}
    \EndFor
\EndFor
\For{\texttt{each $f_i$}}
    \For{\texttt{each $f_j : f_j\neq f_i$}}
        \For{\texttt{each $A^n_i\in \{A^1_i, \dots, A_i^N\}$}}
            \For{\texttt{each $A^m_j\in\{A^1_j,\dots,A^N_j\}$}}
                \State\texttt{Compute $P(F_j\in A^m_j\vert F_i\in A^n_i)$}
            \EndFor
        \EndFor
        \For{\texttt{each $f_k:f_k\neq f_j \neq f_i$}}
            \For{\texttt{each $A_i^n\in \{A^1_i, \dots, A_i^N\}$}}
                \For{\texttt{each $A_j^m\in \{A^1_j, \dots, A_j^N\}$}}
                    \For{\texttt{each $A_k^\ell\in \{A^1_k, \dots, A_k^N\}$}}
                        \State\texttt{Compute $P(F_k\in A_k^\ell\vert F_j\in A_j^m \text{ and } F_i\in A_i^n)$}
                    \EndFor
                \EndFor
            \EndFor
        \EndFor
    \EndFor
\EndFor
\end{algorithmic}
\end{algorithm}

\begin{algorithm}[H]

\caption{Implementation: synthetic data generation using the first order conditional probabilities} \label{alg:gen1}

\begin{algorithmic}[1]
\State\texttt{Let $s=0$ be the row-index of the synthetic dataset}
\State\texttt{Let $S$ be the number of rows in the synthetic dataset}
\While{\texttt{$s<S$}}
    \State\texttt{Select a feature $f_i$, randomly}
    \State\texttt{Select an interval, $A_i^n$, using $\hat{p}_{i,n}$}
    \State\texttt{Draw a value, $x_s$, from $\mathcal{U}(A_i^n)$}
    \For{\texttt{each $f_j : f_j\neq f_i$}}
        \State\texttt{Select an interval, $A_j^m$, using $\hat{p}_{[(j,m)|(i,n)]}$}
        \State\texttt{Draw a value, $y_s$, from $\mathcal{U}(A_j^m)$}
    \EndFor
    \State\texttt{Set $s=s+1$}
\EndWhile
\end{algorithmic}
\end{algorithm}

\begin{algorithm}[H]

\caption{Implementation: synthetic data generation using the second order conditional probabilities} \label{alg:gen2}

\begin{algorithmic}[1]
\State\texttt{Let $s=0$ be the row-index of the synthetic dataset}
\State\texttt{Let $S$ be the number of rows in the synthetic dataset}
\While{\texttt{$s<S$}}
    \State\texttt{Select a feature $f_i$, randomly}
    \State\texttt{Select an interval, $A_i^n$, using $\hat{p}_{i,n}$}
    \State\texttt{Draw a value, $x_s$, from $\mathcal{U}(A_i^n)$}
    \State\texttt{Select a feature $f_j \neq f_i$, randomly}
    \State\texttt{Select an interval, $A_j^m$, using $\hat{p}_{[(j,m)|(i,n)]}$}
    \State\texttt{Draw a value, $y_s$, from $\mathcal{U}(A_j^m)$}
    \For{\texttt{each $f_k : f_k\neq f_i \neq f_j$}}
        \State\texttt{Select an interval, $A_k^\ell$, using $\hat{p}_{[(k,\ell)|(i,n),(j,m)]}$}
        \State\texttt{Draw a value, $z_s$, from $\mathcal{U}(A_k^\ell)$}
    \EndFor
    \State\texttt{Update the matrix with the new observation}
    \State\texttt{Set $s=s+1$}
\EndWhile
\end{algorithmic}
\end{algorithm}

\noindent When generating synthetic data through Algorithms \ref{alg:map1}, \ref{alg:map2}, \ref{alg:gen1} and \ref{alg:gen2} we have made a number of choices. The most important ones refer to the choice of discretization of the feature, the depth of estimated conditional probabilities and how we to pick root-features\footnote{By root-feature we mean the first features in each observation that are used as conditions for the rest of the features in an observation.}. \NJ{A few remarks are warranted regarding the \emph{depth} of conditional dependencies in our generative process. To fully preserve the joint correlations among all features, one would ideally condition each newly generated feature on all previously generated features and their associated intervals. However, this is not the approach taken in Algorithm \ref{alg:gen1} or \ref{alg:gen2}. Instead, we truncate the conditioning at a fixed depth—typically 1 or 2—meaning that each feature is generated conditional only on one or two preceding features. This truncation serves as a tunable parameter that controls the trade-off between fidelity (i.e., preservation of statistical relationships) and confidentiality (i.e., reducing the risk of disclosing sensitive structure). In doing so, the user can adjust the level of correlation preservation according to the privacy requirements of the synthetic data application.}

Since all the computations within the algorithms are completely independent of each other, this method is fully parallelizable. This allows the handling of large original datasets.

\section{Results - comparisons between original and synthetic datasets}\label{current_results}
This section contains two applications of synthetic data generation for different datasets, one manufactured and one real dataset. The evaluation of the method is based on how well we retain the inter feature correlations of the original dataset in the synthetic once. The measure of correlation we use is the Pearson correlation coefficient between all features of a dataset, leading to a matrix, say $C$. If we take two features $f_j$ and $f_j$ of a dataset $\mathcal{O}$, the corresponding entry in the Pearson correlation matrix is given by

\begin{equation}
    C_{ij} := \frac{\text{cov}(f_i, f_j)}{\sigma_{f_i}\sigma_{f_j}},
    \label{eq:PearsonCorr}
\end{equation}
where $\sigma_{f_i}$ is the standard deviation of feature $f_i$. The covariance of two features is calculated in the following way
\begin{equation}
    \text{cov}(f_i, f_j) := \mathbbl{E}[(f_i - \mu_{f_i})(f_j - \mu_{f_j})],
\end{equation}
where $\mu_{f_i}$ is the mean value of $f_i$ and $\mathbbl{E}[\cdot]$ is the expected value, for more details see \cite{Casella2024} or any textbook on statistics/probability.

\subsection{Application and evaluation on a manufactured example}\label{sec:computationalExample}
\NJ{To evaluate the methodology, we first use a manufactured dataset where the relationships between features are known. 
Let $X$ and $Y$ be independent standard normal random variables i.e.}
\begin{align*}
    X\sim\mathcal{N}(0,1),\qquad Y\sim\mathcal{N}(0,1),\qquad X\perp Y.
\end{align*}
\NJ{We introduce a set of random variables $F_1, \dots, F_6$ via the following\\ equations:}
\begin{align*}
    F_1 &= X, \\
    F_2 &= Y, \\
    F_3 &= 2X + \varepsilon_1,\\
    F_4 &= \sin({X}) + \varepsilon_2,\\
    F_5 &= \log(|X| + 1) + \varepsilon_3, \\
    F_6 &= \frac{1}{2}X + \frac{1}{2}Y + \varepsilon_4.
\end{align*}
\NJ{The noise terms $\varepsilon_j$ are zero-mean Gaussian variables with standard deviation chosen relative to the standard deviation of the deterministic part of each feature as in }
\begin{equation}
    \varepsilon_j \sim \mathcal{N}(0, \sigma_j^2), \quad\text{where  } \sigma_j = \alpha  \sigma_{f_j},
\end{equation}
\NJ{and $f_j$ is the noiseless component of feature $F_j$, and $\alpha\in [0, 1]$ is a noise scaling parameter controlling the signal-to-noise ratio, where $\sigma_{f_j}$ denotes the standard deviation of the variable $f_j$. 
Let $\mathcal{O}\in R^{M\times6}$ be a dataset consisting of $M$ i.i.d. realizations $(f^{(i)}_1,\dots, f^{(i)}_6)\sim(F_1, \dots, F_6)$.}

\NJ{The algorithms described in Section \ref{Implementation} allow for varying levels of conditional probabilities. 
Specifically, one can chose to use the marginal probability estimates $\hat{p}_{i,k}$ in \eqref{eq:measure_mu_i} and the pairwise conditional probabilities $\hat{p}_{[(j,m), (i,n)]}$ from \eqref{eq:emp_condprob}, or extend with the second-order conditional probability estimates $\hat{p}_{[(k,l)|(i,n),(j,m)]}$ in \eqref{eq:emp_condprob_2}.
Including the higher-order conditional structure in \eqref{eq:emp_condprob_2} significantly increases the computational cost. However, it is also evident that it leads to synthetic datasets that more closely replicate the joint dependencies in the original data.}

\begin{figure}
\begin{picture}(400,350)(20,0)
\put(0,260)
{
 \includegraphics[width=.50\textwidth, height=.50\textwidth]{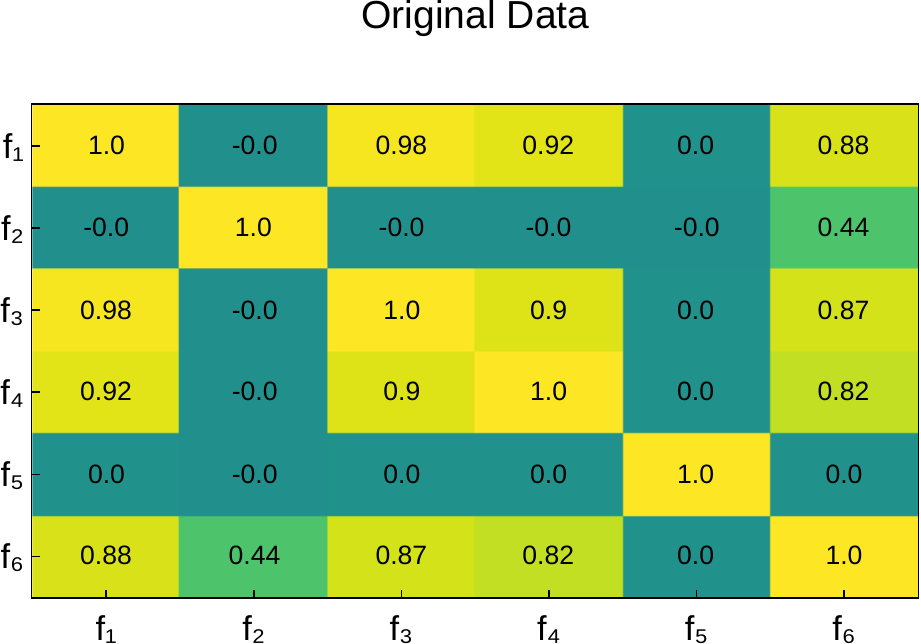}
}
\put(200,260)
{
 \includegraphics[width=.50\textwidth, height=.50\textwidth]{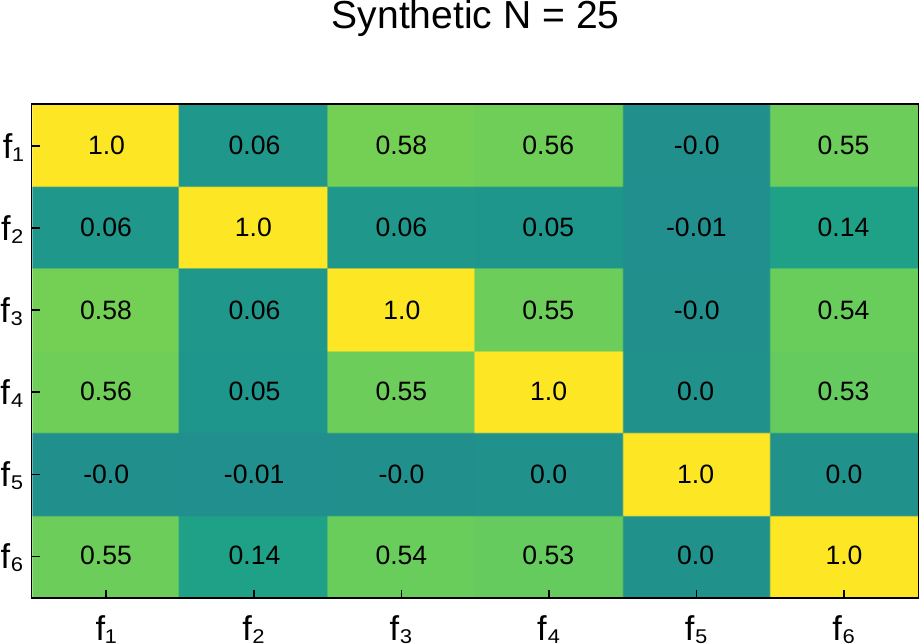}
}
\put(0,130)
{
 \includegraphics[width=.33\textwidth, height=.33\textwidth]{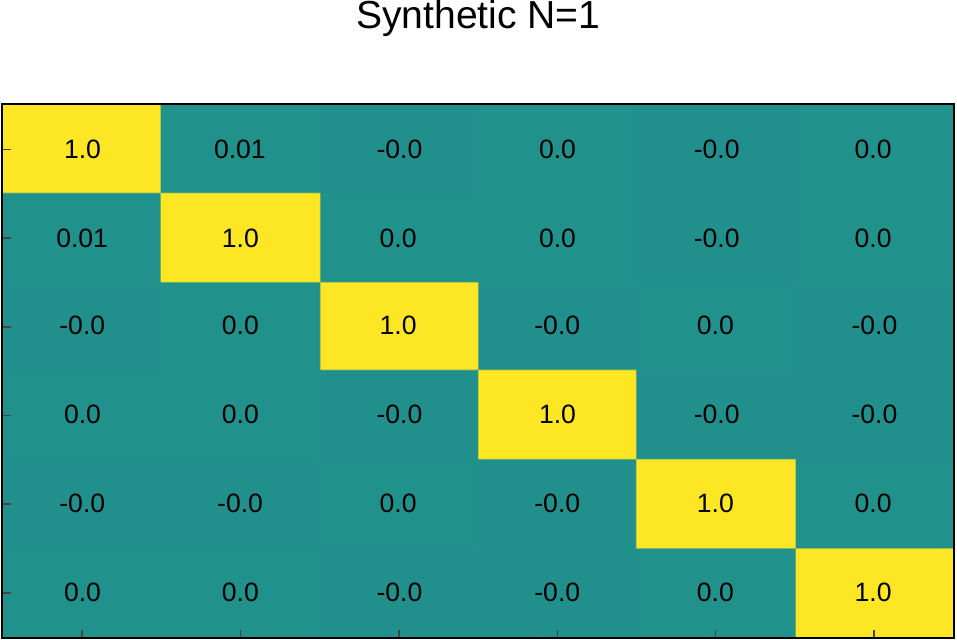}
}
\put(130,130)
{
 \includegraphics[width=.33\textwidth, height=.33\textwidth]{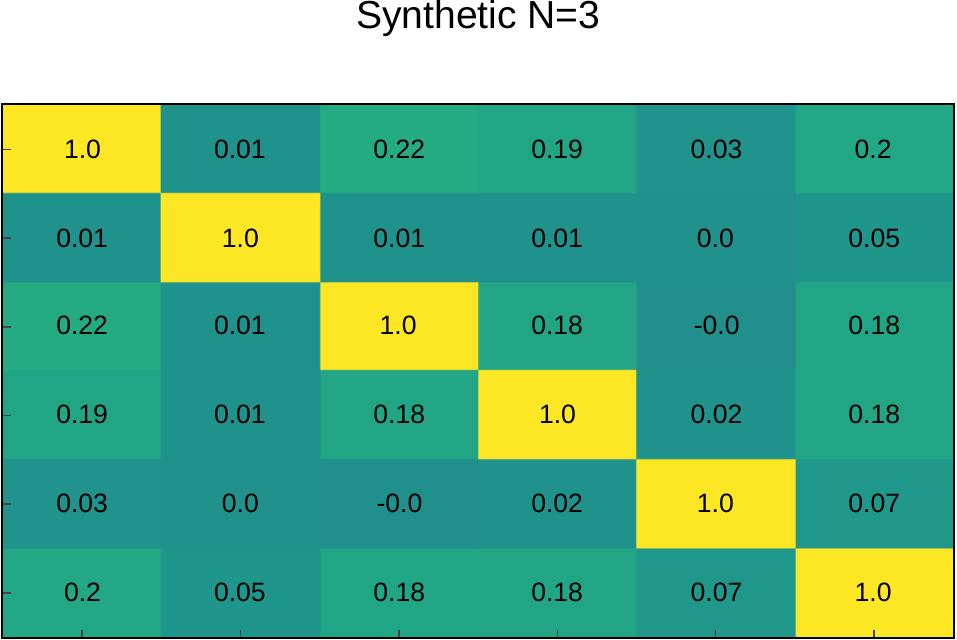}
}
\put(260,130)
{
 \includegraphics[width=.33\textwidth, height=.33\textwidth]{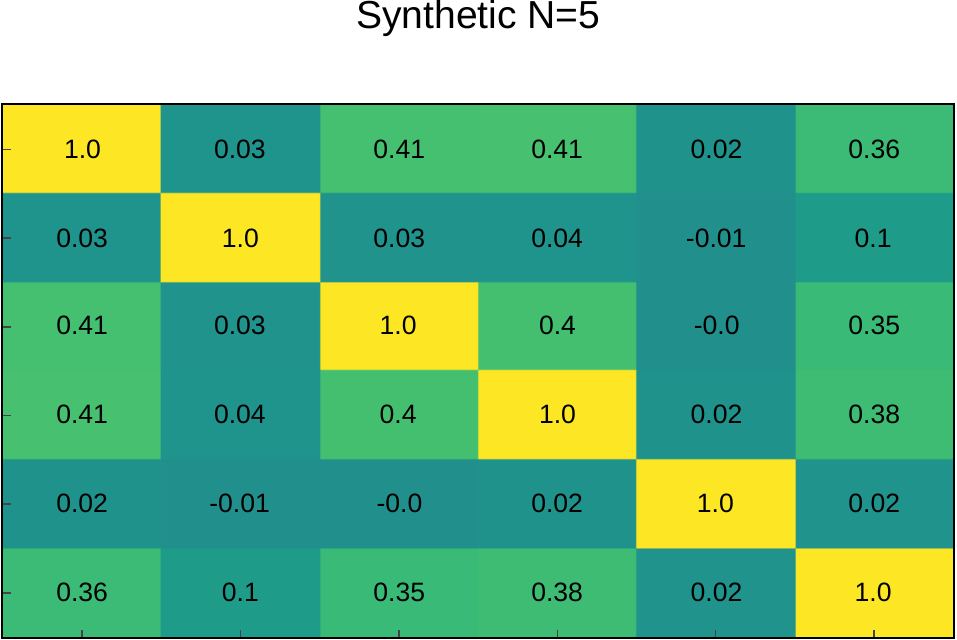}
}
\put(0,0)
{
 \includegraphics[width=.33\textwidth, height=.33\textwidth]{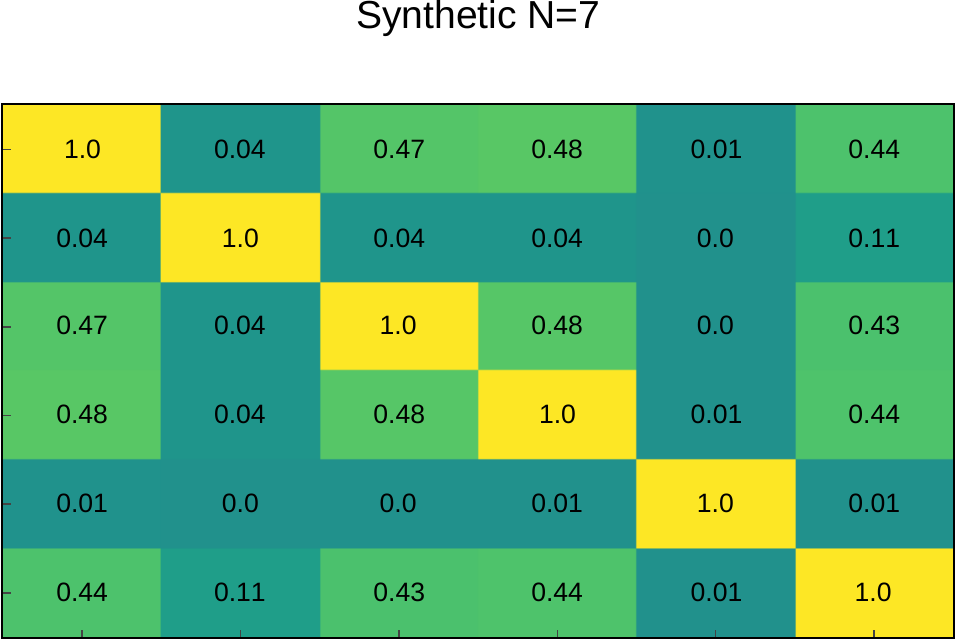}
}
\put(130,0)
{
 \includegraphics[width=.33\textwidth, height=.33\textwidth]{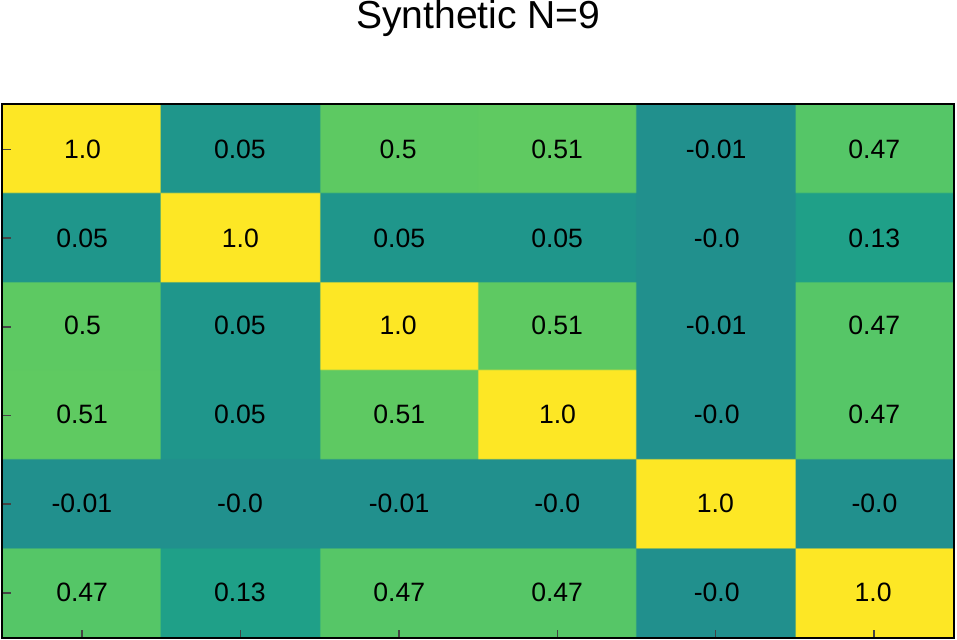}
}
\put(260,0)
{
\includegraphics[width=.33\textwidth, height=.33\textwidth]{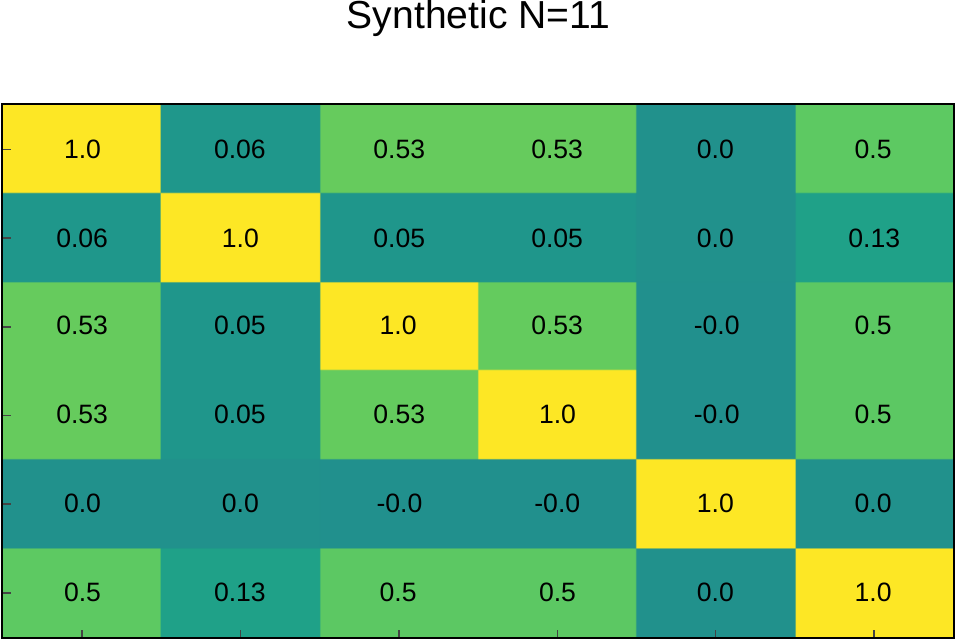}
}
\put(5,-25)
{
\includegraphics[width=\textwidth, height=0.05\textwidth]{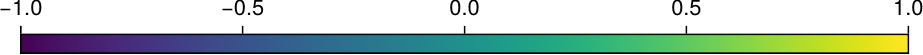}
}
\end{picture}
\caption{
Top panel: Pearson matrix of the original dataset (left) and synthetic dataset (right) computed using first order conditional distribution with $N=25$. Bottom panel: Pearson matrix for synthetic data using first order conditional distributions with increasing $N$. The features in the bottom panel are ordered in the same manner as is displayed in the top panel. 
}
\label{fig:Example_corr_Order1}
\end{figure}

\NJ{In Figure \ref{fig:Example_corr_Order1}, we compare the Pearson correlation matrices of the original and synthetic datasets, where the synthetic data is generated using only first-order conditional probabilities under varying discretizations. The results show that increasing $N$ improves the alignment of feature correlations with the original data.}

\begin{figure}
\begin{picture}(400,350)(20,0)
\put(0,260)
{
 \includegraphics[width=.50\textwidth, height=.50\textwidth]{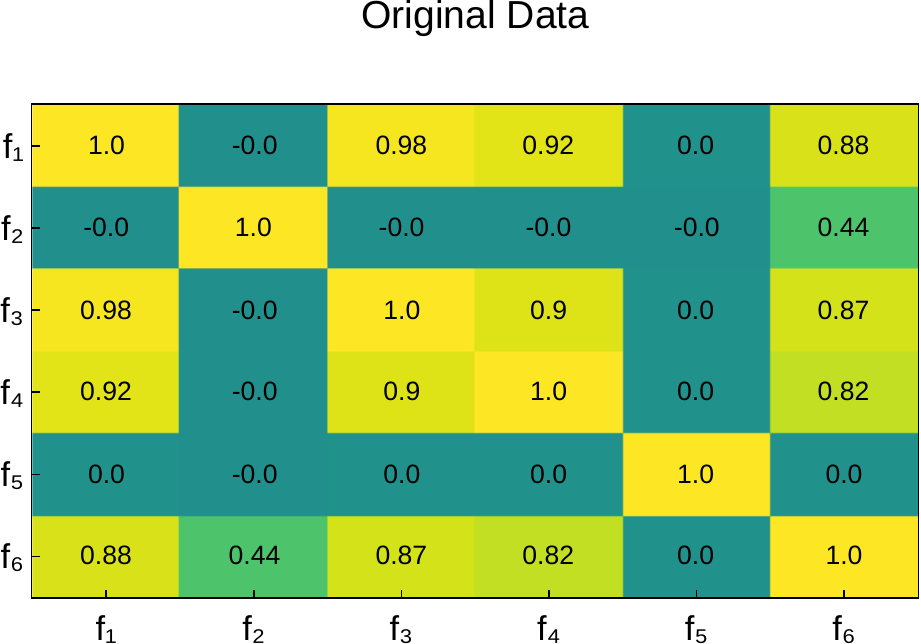}
}
\put(200,260)
{
 \includegraphics[width=.50\textwidth, height=.50\textwidth]{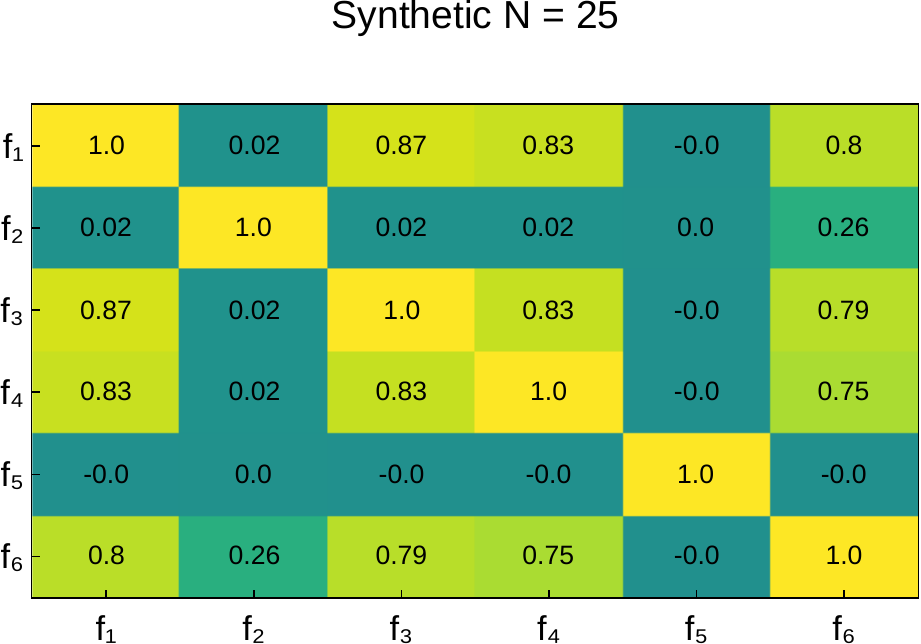}
}
\put(0,130)
{
 \includegraphics[width=.33\textwidth, height=.33\textwidth]{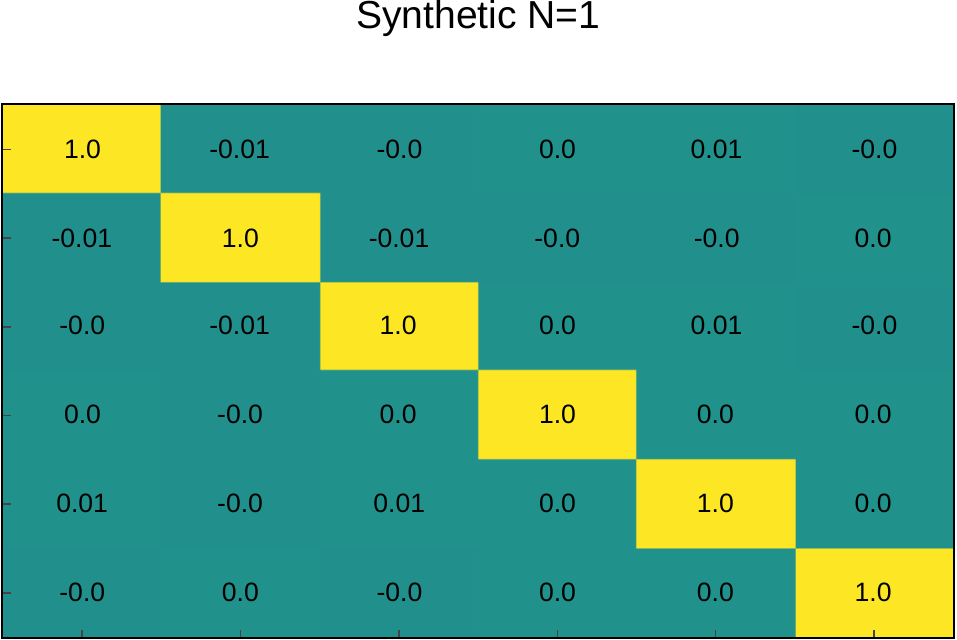}
}
\put(130,130)
{
 \includegraphics[width=.33\textwidth, height=.33\textwidth]{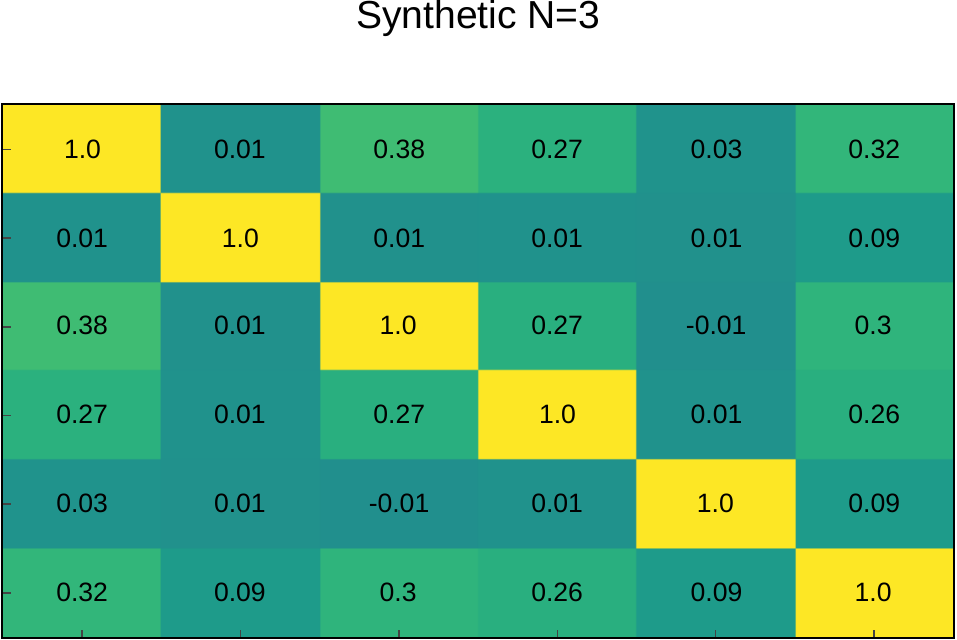}
}
\put(260,130)
{
 \includegraphics[width=.33\textwidth, height=.33\textwidth]{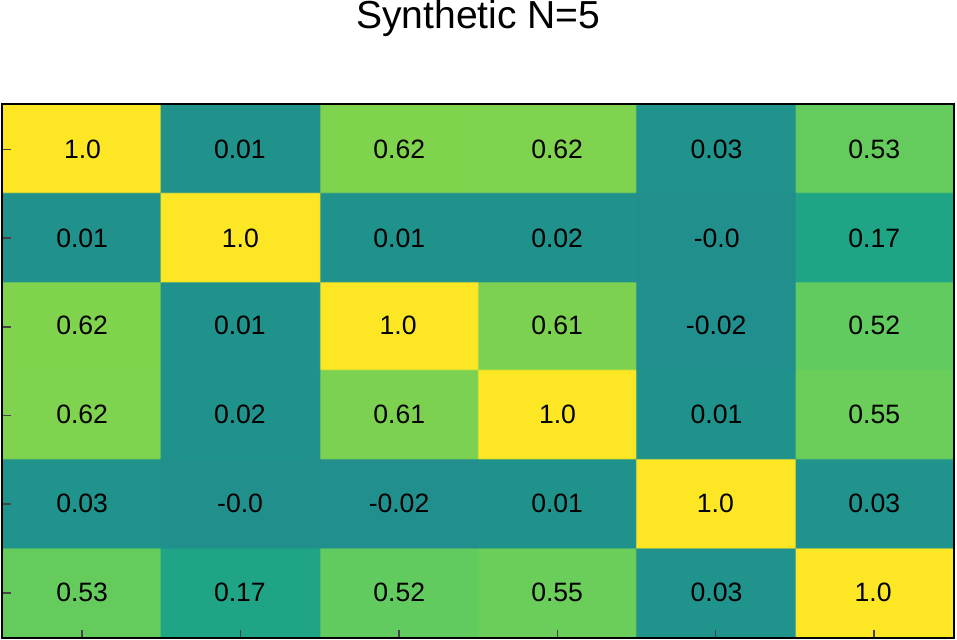}
}
\put(0,0)
{
 \includegraphics[width=.33\textwidth, height=.33\textwidth]{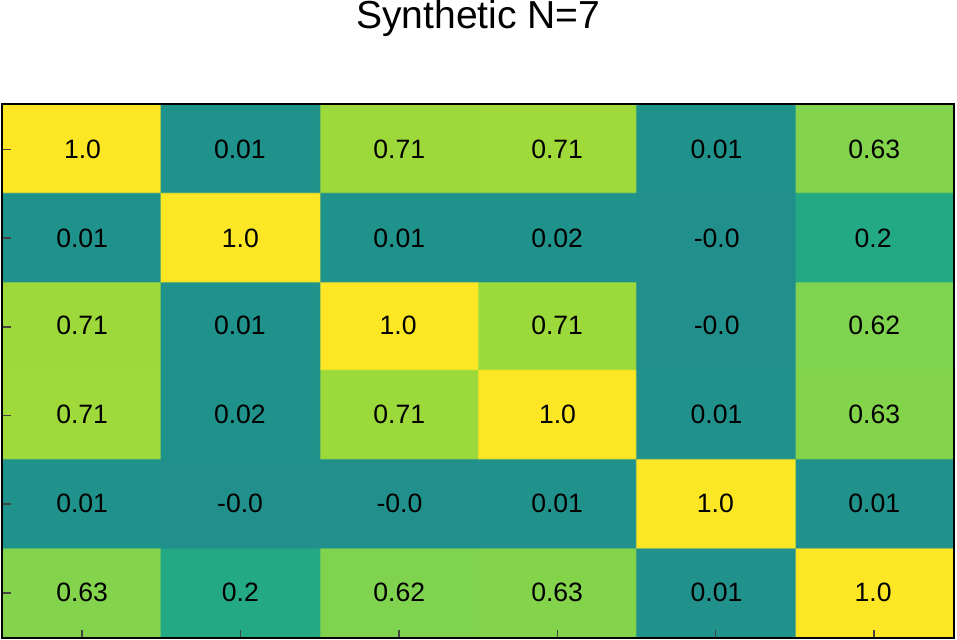}
}
\put(130,0)
{
 \includegraphics[width=.33\textwidth, height=.33\textwidth]{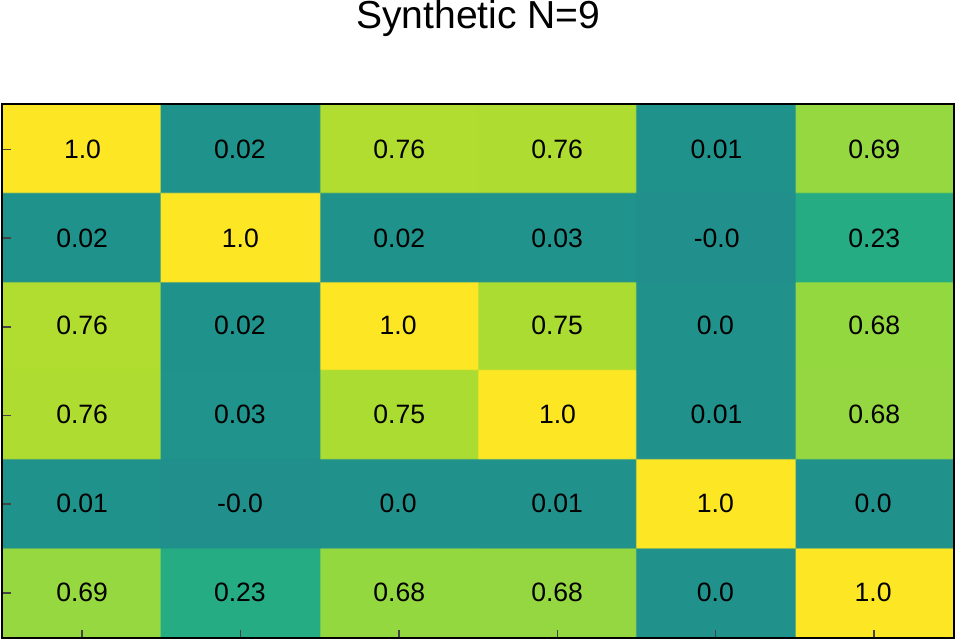}
}
\put(260,0)
{
\includegraphics[width=.33\textwidth, height=.33\textwidth]{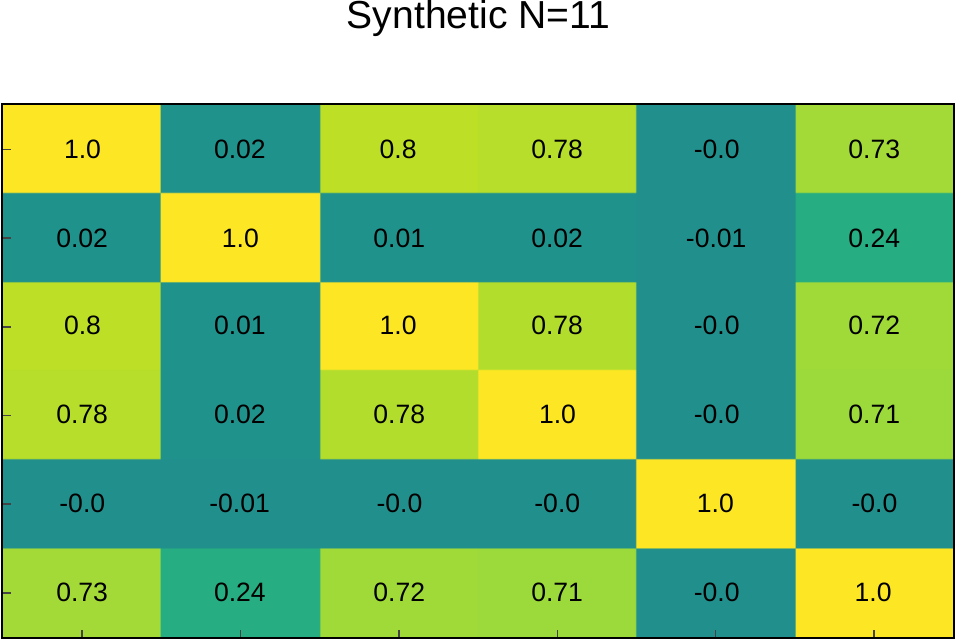}
}
\put(5,-25)
{
\includegraphics[width=\textwidth, height=0.05\textwidth]{colorbar.pdf}
}
\end{picture}
\caption{
Top panel: Pearson matrix of the original dataset (left) and synthetic dataset (right) computed using first and second order conditional distribution with $N=25$. Bottom panel: Pearson matrix for synthetic data using first and second order conditional distributions with increasing $N$. The features in the bottom panel are ordered in the same manner as is displayed in the top panel. 
}
\label{fig:Example_corr_Order2}
\end{figure}

\NJ{Figure \ref{fig:Example_corr_Order2} presents results obtained using second-order conditional probabilities. These demonstrate that incorporating higher-order dependencies leads to a closer preservation of the correlation structure, even at coarser discretization. Overall, the results highlight a few things. There is a tradeoff between computational efficiency and correlational fidelity. It also begins to show how the depth or correlation could be used as a parameter to control how much information is being transferred from the original to the synthetic data.}

\subsection{Application to a real dataset}
In what follows we apply the proposed methodology to the dataset presented in section \ref{acquisition}.

\subsection{Data acquisition}\label{acquisition}

The dataset was attained from \cite{pereira_sustdata}. The dataset contains over 35 million individual records of electric energy related data, among which you can find consumption and demographic information from 50 monitored homes together with electric energy production in Madeira Island and supporting environmental data. 
\texttt{SustData} has been used in the research on Non-Intrusive Load Monitoring (NILM)\cite{Renaux2020} and particularly in event based approaches for NILM as discussed in \cite{Ruano2019NILMTF}. 
The subset we work on is home power consumption data containing 15 features (columns) representing the minimum, maximum, and average of current ($I$), voltage ($V$), real power ($P$), power factor ($PF$), and reactive power ($Q$) with a temporal resolution of one minute. A more detailed description of the dataset is available in \cite{Pereira2014}. 
From this dataset we extracted a random sample with circa $10$ million individual records. This subset was then cleaned, removing any rows of observations that had values missing or the like. 
Also any categorical columns were removed, since our approach, for now, only deals with floating point numbers. 
Any rows containing invalid entries were also removed. From this cleaned version of the dataset, we sample randomly $5$ million observations. We refer to the reduced (clean) dataset as $\mathcal{O}$.
Its features, will  be denoted by $f_i (i\in\{1,\dots,15\})$. What concerns the dataset used within this framework, we identify the features as follows in Table \ref{tab:Dataset}.

\begin{table}[ht]
\centering
\begin{tabular}{ccc}
 Notation& Physical meaning& Symbol\\
 \cmidrule{1-3}
 $f_{1}$& Minimum Current & \hspace{-1em}\rdelim\}{3}{*}[I]\\
 $f_{2}$& Maximum Current\\
 $f_{3}$& Average Current\\
 $f_{4}$& Minimum Voltage & \hspace{-1em}\rdelim\}{3}{*}[V]\\
 $f_{5}$& Maximum Voltage\\
 $f_{6}$& Average Voltage\\
 $f_{7}$& Minimum Power & \hspace{-1em}\rdelim\}{3}{*}[P]\\
 $f_{8}$& Maximum Power\\
 $f_{9}$& Average Power\\
 $f_{10}$& Minimum Power Factor &\hspace{-1em}\rdelim\}{3}{*}[PF]\\
 $f_{11}$& Maximum Power Factor\\
 $f_{12}$& Average Power Factor\\
 $f_{13}$& Minimum Reactive Power&\hspace{-1em}\rdelim\}{3}{*}[Q]\\
 $f_{14}$& Maximum Reactive Power\\
 $f_{15}$& Average Reactive Power
\end{tabular}
\caption{Description of the notation and physical meaning of each feature of the dataset $\mathcal{O}$. The maximum, minimum and average are taken over the course of one minute.}
\label{tab:Dataset}
\end{table}

\subsubsection{Comparison of first-order distributions}

\begin{figure}
    \centering
    \includegraphics[width=12cm]{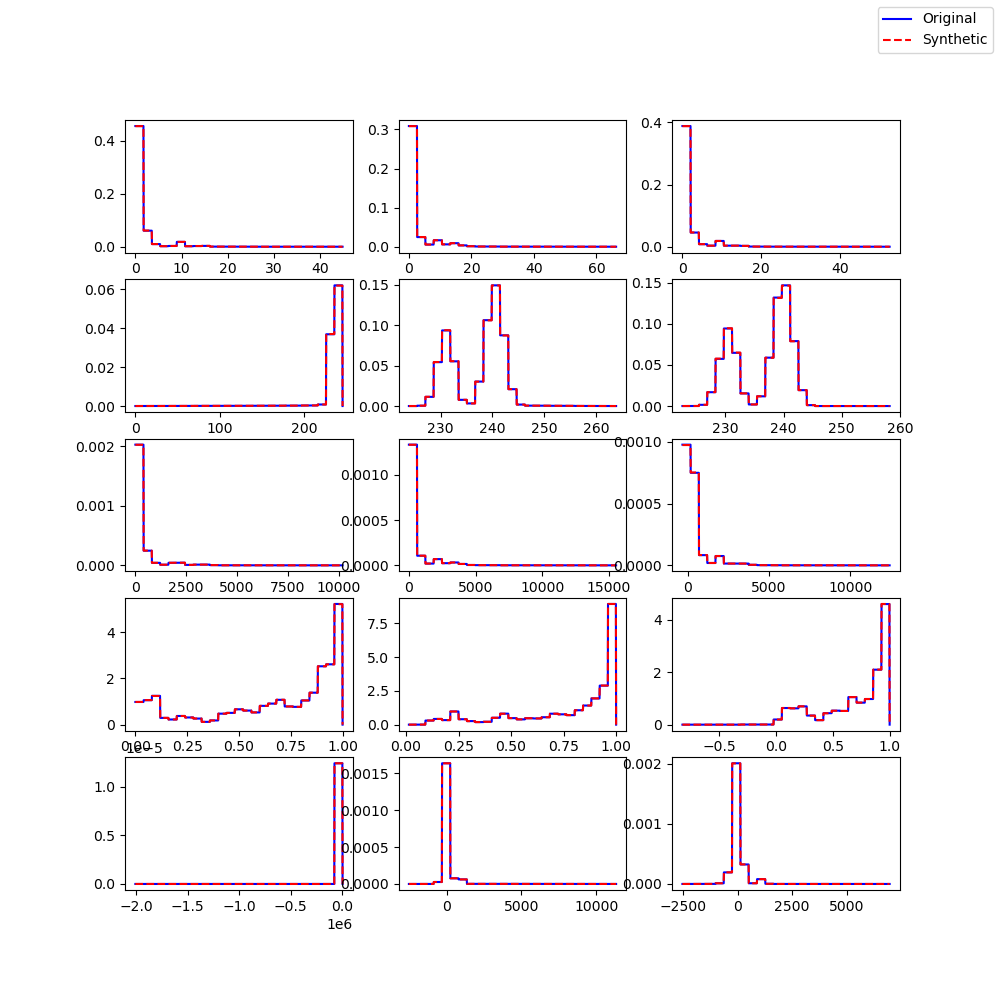}
    \caption{Comparison of the distributions between then synthetic (red) and original (blue) data using $N=25$ intervals. The subplots (left to right, up to down) are the features of the dataset in increasing order.}
    \label{fig:distcomp}
\end{figure}

\noindent In Figure \ref{fig:distcomp}, we plot the distributions of each feature from the original ($\mathcal{O}$) and the synthetic ($\mathcal{S}$) dataset. We illustrate herewith the similarity between the distributions of both datasets.
The observed similarity in Figure \ref{fig:distcomp} is very good, as expected. This is essentially showing that the synthetic dataset is a well sampled representation of the original dataset at this resolution ($N=25$). 

\subsubsection{Comparison of second-order distributions}

\begin{figure}
    \centering
    \includegraphics[width=12cm]{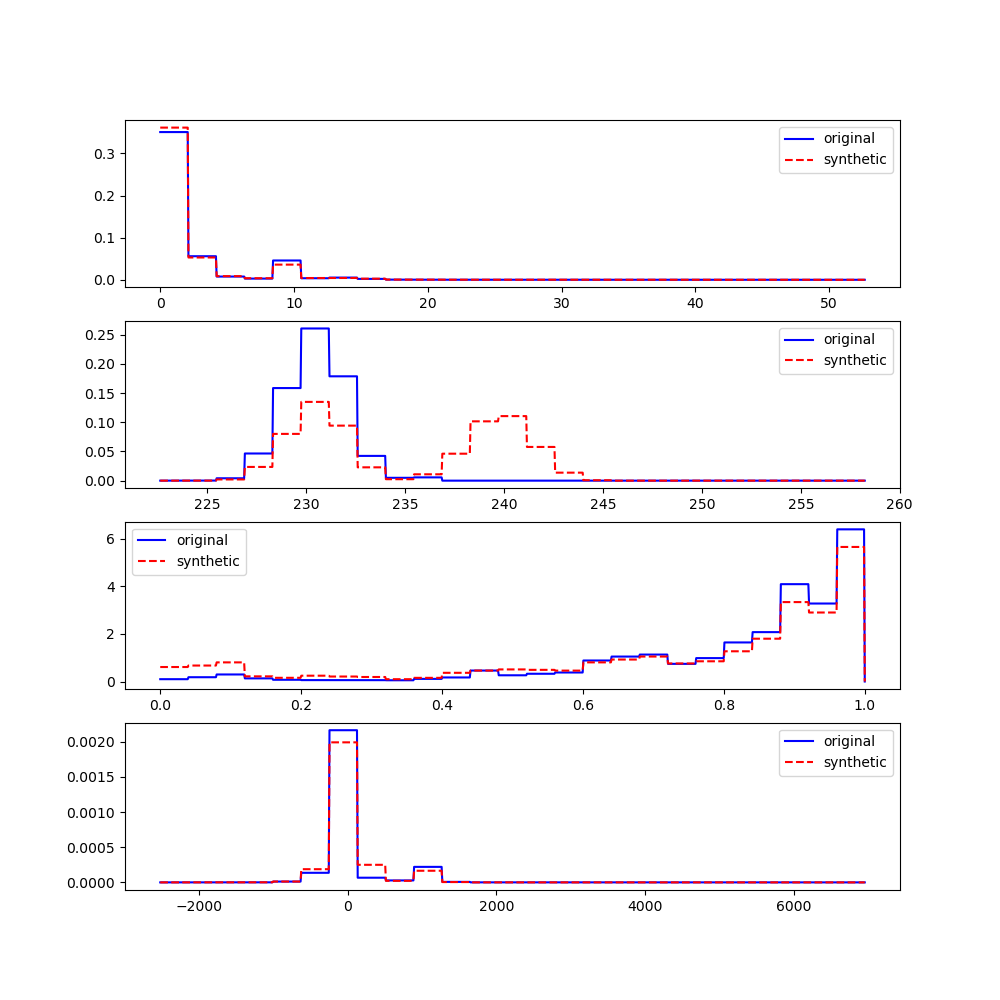}
    \caption{Comparison between the conditional distributions of features 3, 6, 10, and 15 of the original (blue) and the synthetic (red) dataset given that the corresponding values in feature 5 are in the first third of its range. The synthetic dataset was computed using only first-order conditional probabilities with $N=25$ intervals of discretization.}
    \label{fig:condDistEx750}
\end{figure}

To illustrate the retention of correlations between features of $\mathcal{S}$ compared to $\mathcal{O}$, we select one feature, $f_i$, and one set in the range of $f_i$, called generically $A$. 
In Figure \ref{fig:condDistEx750} we show the conditional distribution of features 3, 6, 10, and 15 given that the corresponding values in feature 5 are from $A$, while $A$ is chosen to be the first third of the range. \NJ{These choices are to illustrate what kind of things can go wrong whilst simultaneously give a general estimate of how close they can look.}
The more similar the correlations within $\mathcal{O}$ and $\mathcal{S}$ are, the more similar we would expect their conditional distributions to be. 

\begin{figure}
    \centering
    \includegraphics[width=12cm]{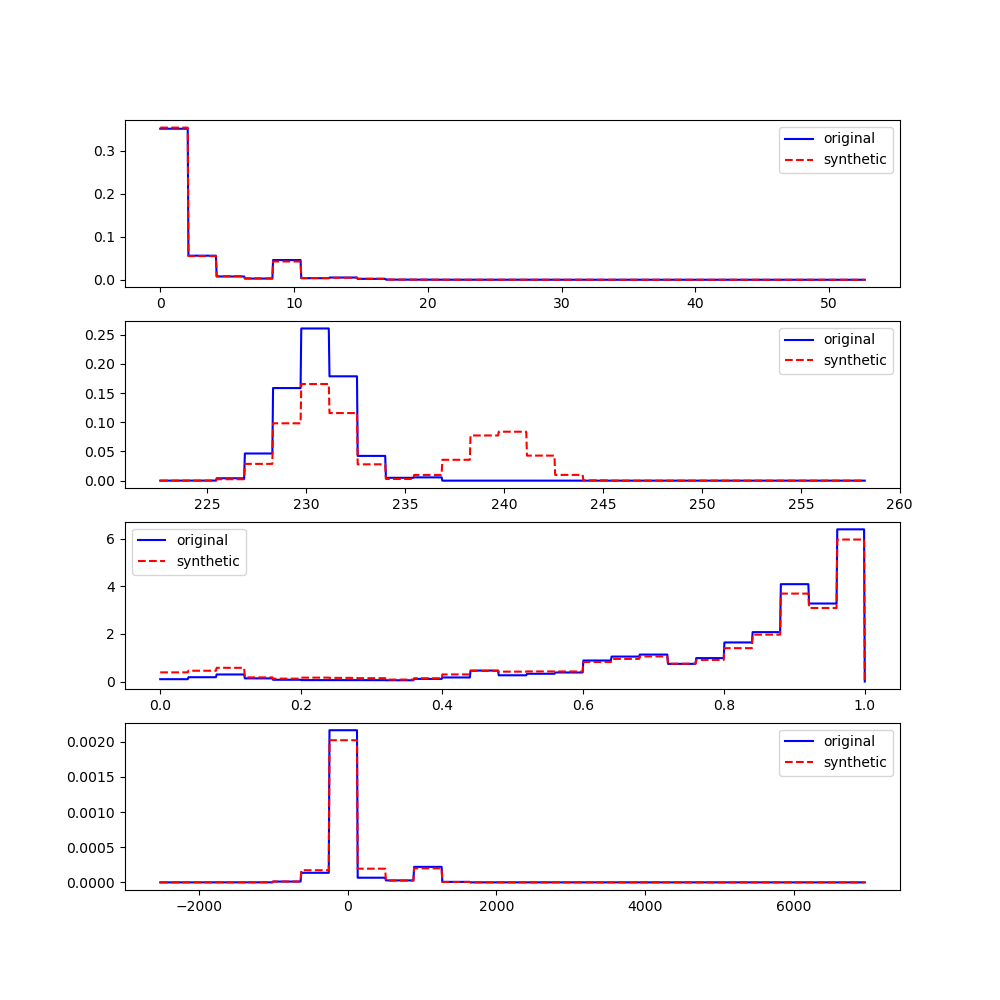}
    \caption{Comparison between the second-order conditional distributions of features 3, 6, 10, and 15 of the original (blue) and the synthetic (red) dataset given that the corresponding values in feature 5 are in the first third of its range. $N=25$}
    \label{fig:condDistOrder2}
\end{figure}

Referring to Figure \ref{fig:condDistEx750}, particularly in the second plot it is evident that the conditional probabilities of the synthetic data \NJ{mistakenly} retain the bimodality displayed by feature 6 in Figure \ref{fig:distcomp}. \NJ{This is due to the fact that some features are independent of feature 6 such that if one such feature is chosen to be the root feature, that conditional distribution would display bimodality.}
Turning our attention to Figure \ref{fig:condDistOrder2}, we see the same type of plot as Figure \ref{fig:condDistEx750} but the synthetic dataset was computed using the second-order conditional distributions. 
Visually, the mismatch between the original and synthetic dataset of Figure \ref{fig:condDistOrder2} is everywhere smaller than in Figure \ref{fig:condDistEx750}.
This aspect is most clearly shown in the second plot of Figure \ref{fig:condDistOrder2}. \NJ{This can be understood by realizing that the number of paths through the condition tree that mistakenly retains the bimodality of feature 6 has been reduced by the additional conditions.}
\noindent Note that it is expected that the difference between the datasets $\mathcal{S}$ and $\mathcal{O}$ depends on both the order of conditional probabilities used in generation and the corresponding choice of discretization. 
When comparing Figure \ref{fig:condDistEx750} and Figure \ref{fig:condDistOrder2} it is clear that the order of conditional probabilities matters.
It is also clear that if we wish to represent perfectly the correlations of the original dataset, second-order conditional probabilities are not enough. If a better representation is necessary, then higher order conditional probabilities should be used.

\begin{figure}
\begin{picture}(400,350)(20,0)
\put(-10,260)
{
 \includegraphics[width=.50\textwidth, height=.50\textwidth]{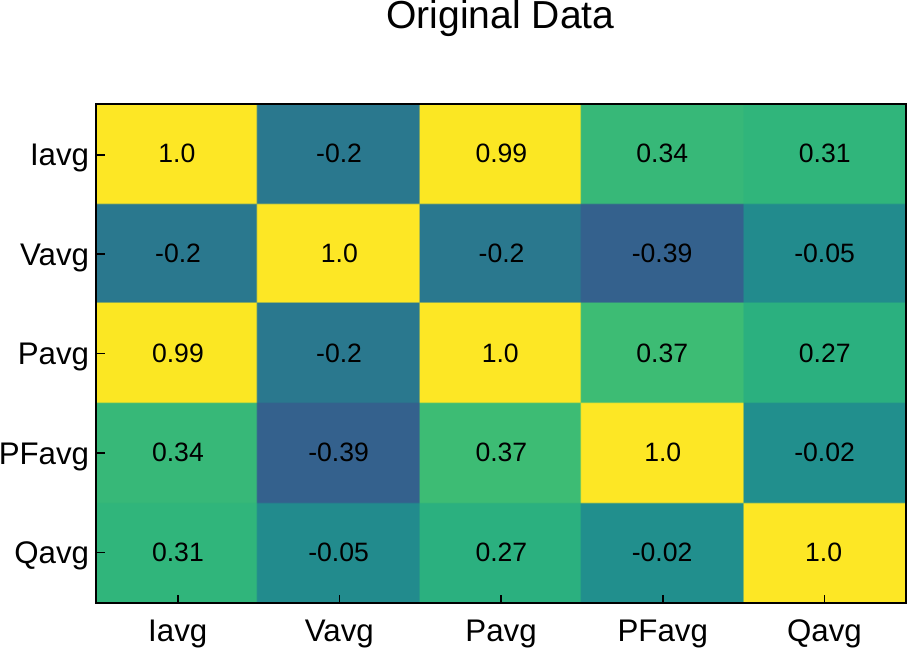}
}
\put(190,260)
{
 \includegraphics[width=.50\textwidth, height=.50\textwidth]{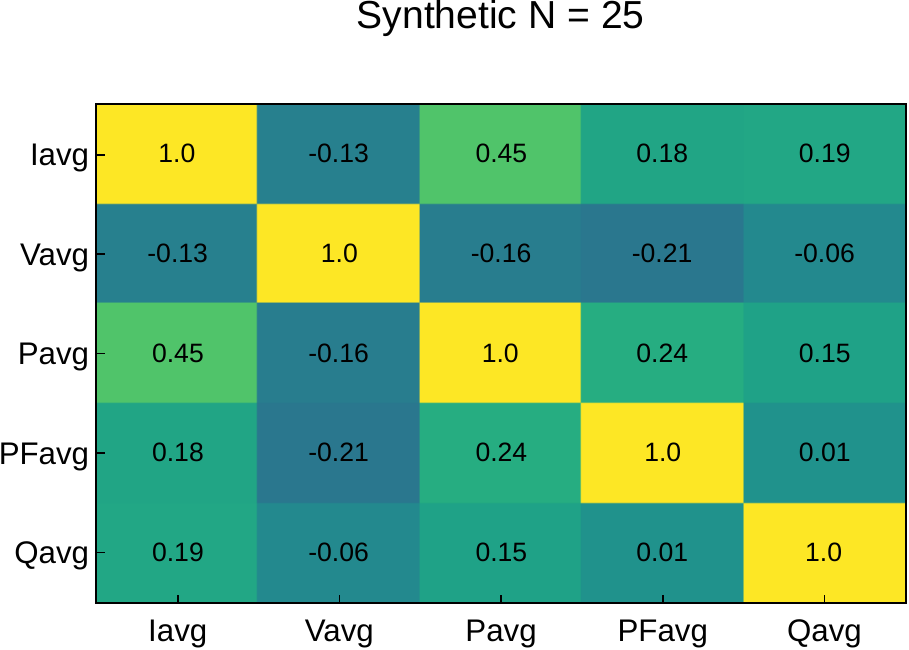}
}
\put(0,130)
{
 \includegraphics[width=.33\textwidth, height=.33\textwidth]{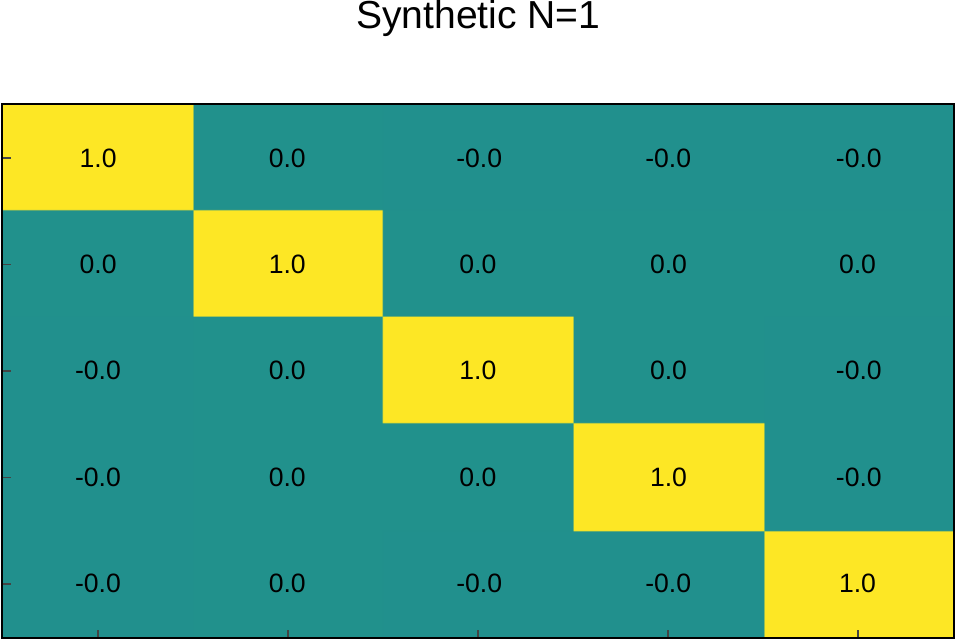}
}
\put(130,130)
{
 \includegraphics[width=.33\textwidth, height=.33\textwidth]{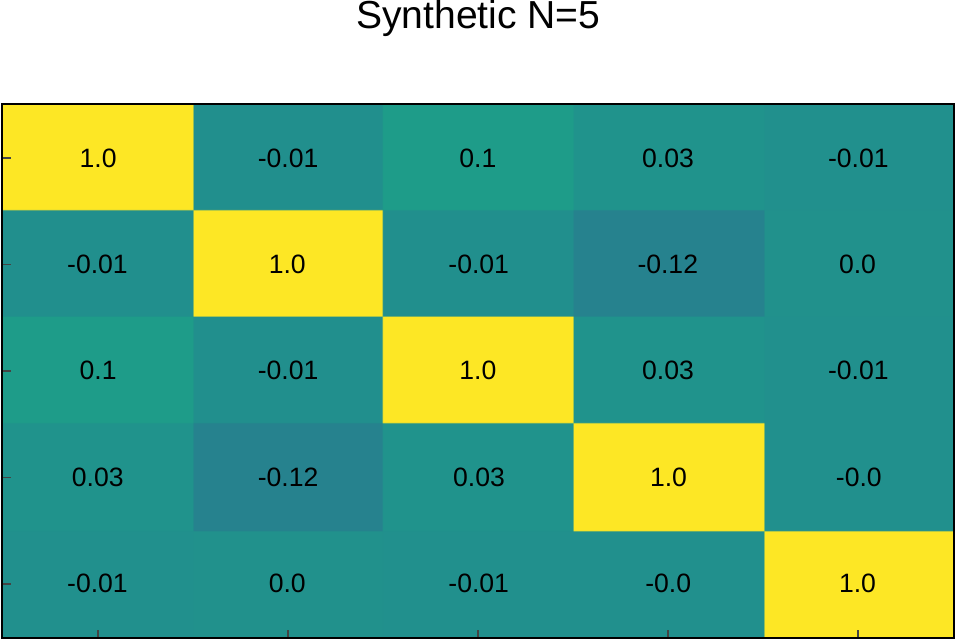}
}
\put(260,130)
{
 \includegraphics[width=.33\textwidth, height=.33\textwidth]{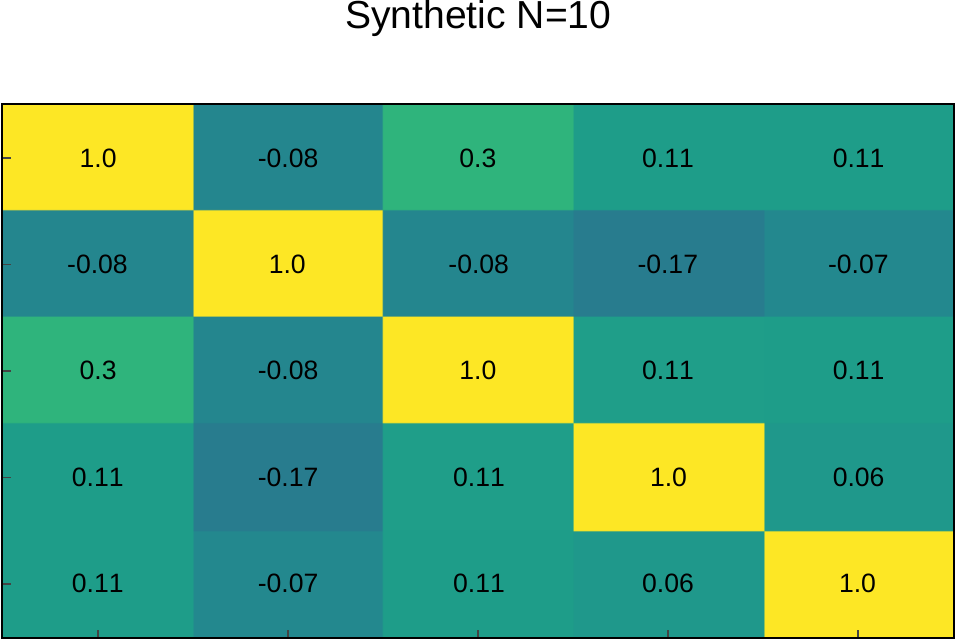}
}
\put(0,0)
{
 \includegraphics[width=.33\textwidth, height=.33\textwidth]{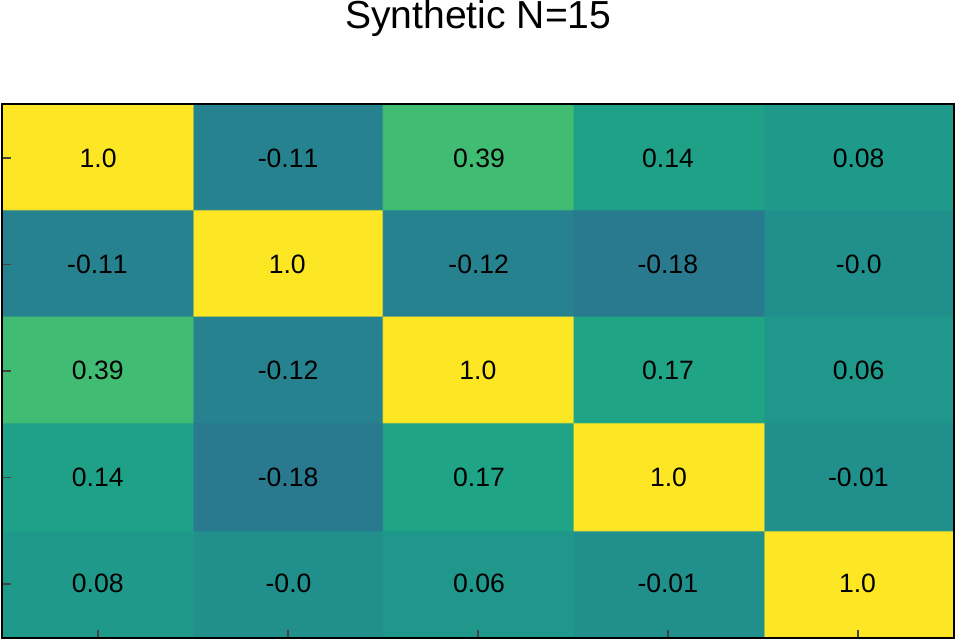}
}
\put(130,0)
{
 \includegraphics[width=.33\textwidth, height=.33\textwidth]{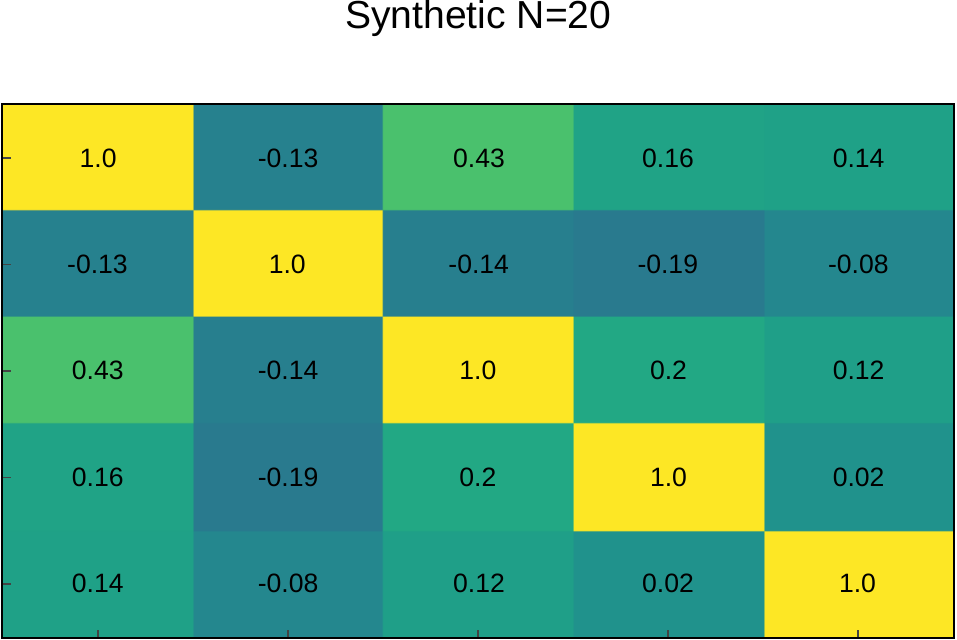}
}
\put(260,0)
{
\includegraphics[width=.33\textwidth, height=.33\textwidth]{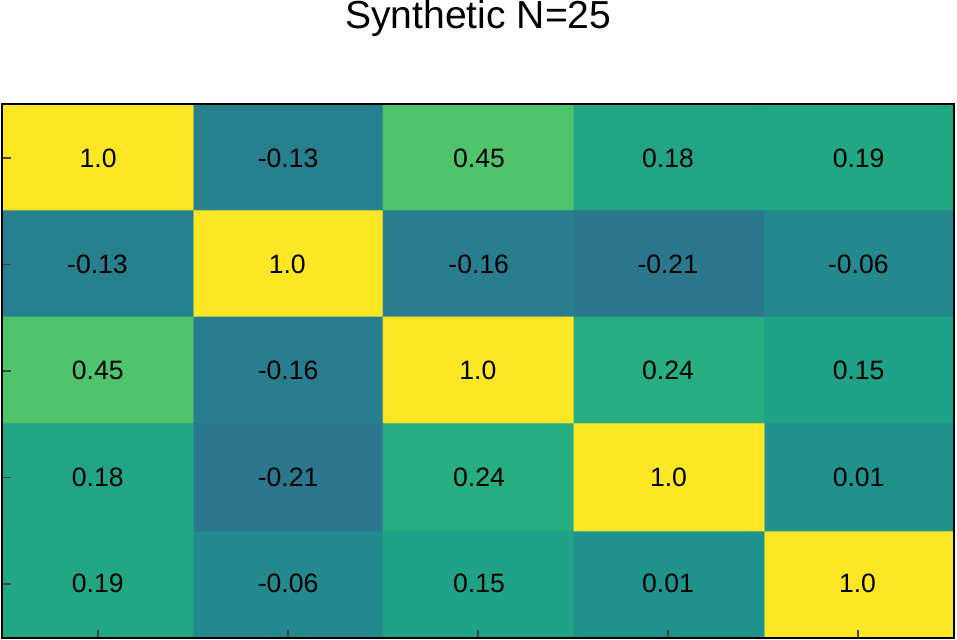}
}
\put(5,-25)
{
\includegraphics[width=\textwidth, height=0.05\textwidth]{colorbar.pdf}
}
\end{picture}
\caption{
Top panel: Pearson matrix of the original (\texttt{SustData}) dataset (left) and synthetic dataset (right) computed using first order conditional distribution with $N=25$. Bottom panel: Pearson matrix for synthetic data using first order conditional distributions with increasing $N$. The features in the bottom panel are ordered in the same manner as is displayed in the top panel. 
}
\label{fig:PearsonCorrOrder1}
\end{figure}

\begin{figure}
\begin{picture}(400,350)(20,0)
\put(-10,260)
{
 \includegraphics[width=.50\textwidth, height=.50\textwidth]{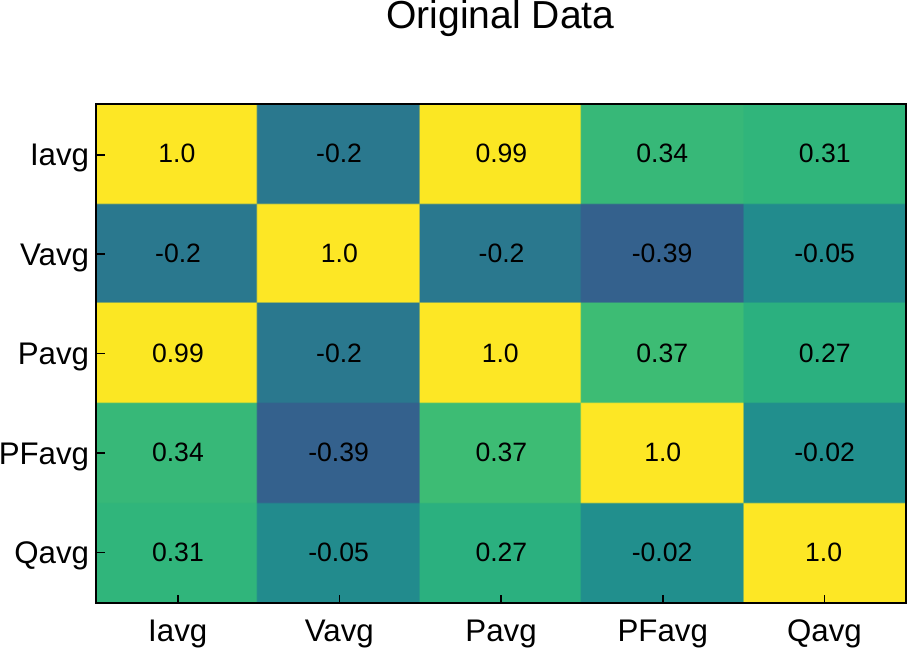}
}
\put(190,260)
{
 \includegraphics[width=.50\textwidth, height=.50\textwidth]{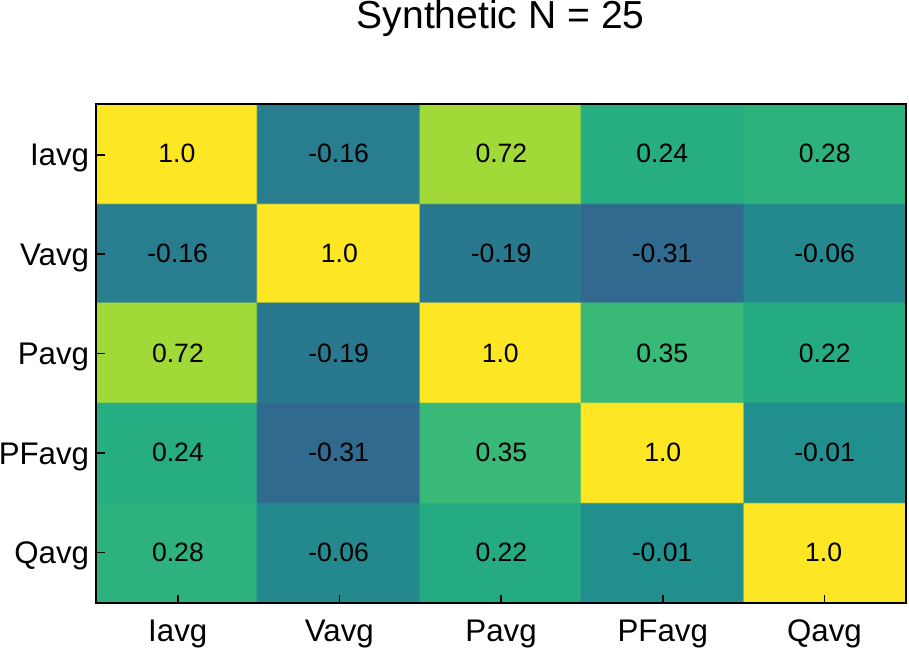}
}
\put(0,130)
{
 \includegraphics[width=.33\textwidth, height=.33\textwidth]{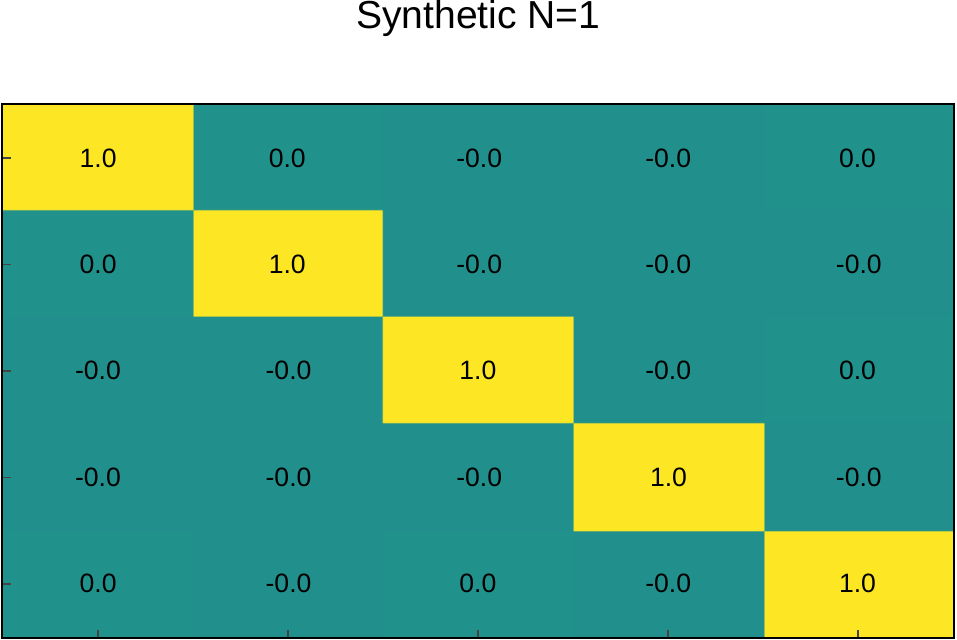}
}
\put(130,130)
{
 \includegraphics[width=.33\textwidth, height=.33\textwidth]{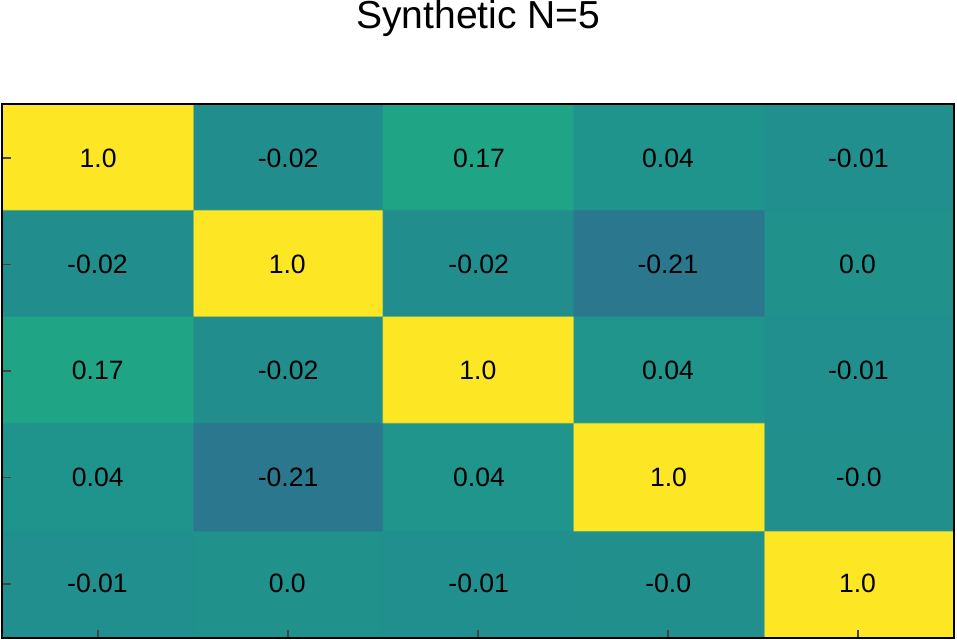}
}
\put(260,130)
{
 \includegraphics[width=.33\textwidth, height=.33\textwidth]{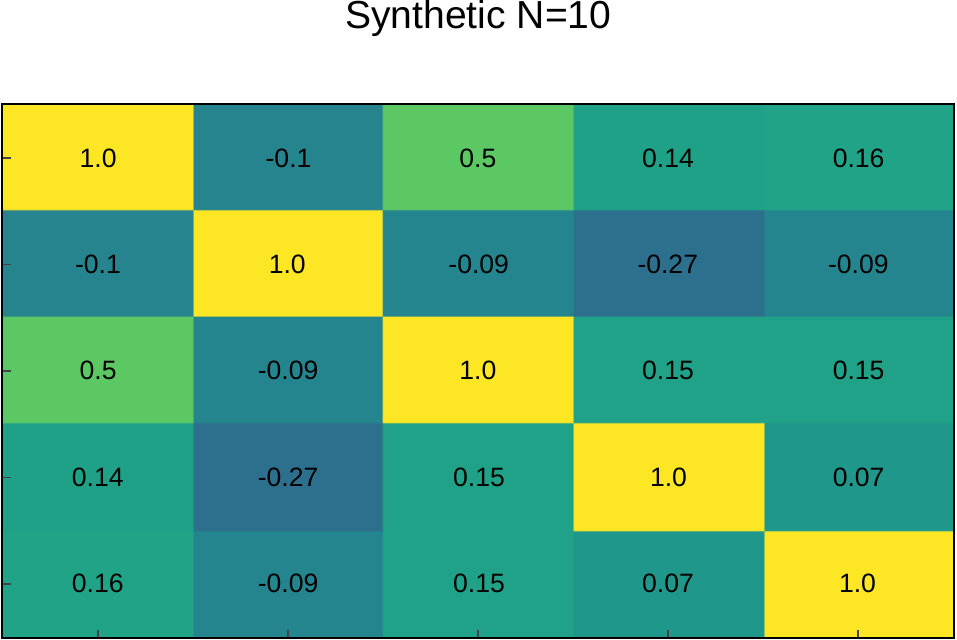}
}
\put(0,0)
{
 \includegraphics[width=.33\textwidth, height=.33\textwidth]{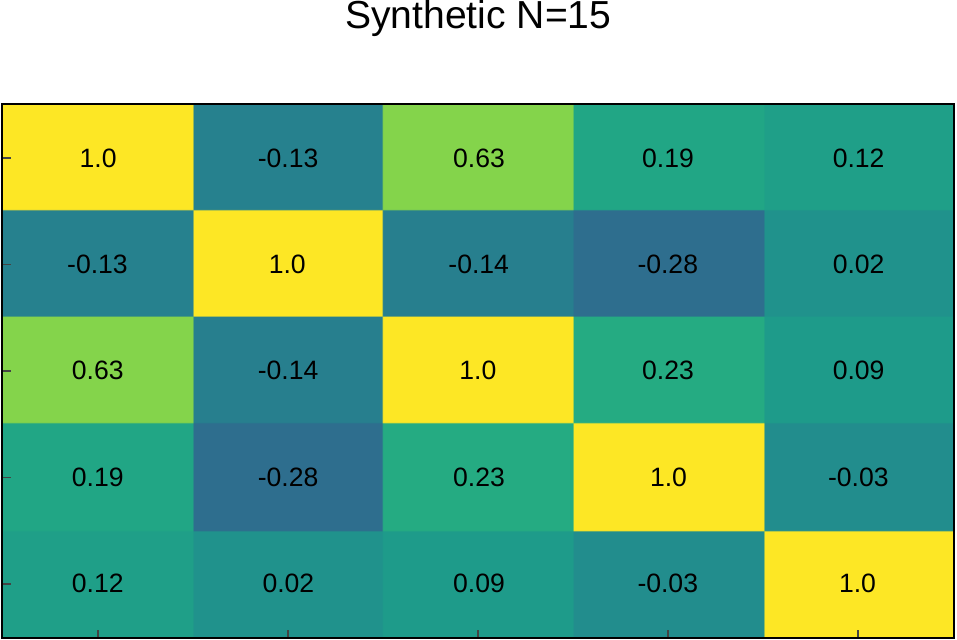}
}
\put(130,0)
{
 \includegraphics[width=.33\textwidth, height=.33\textwidth]{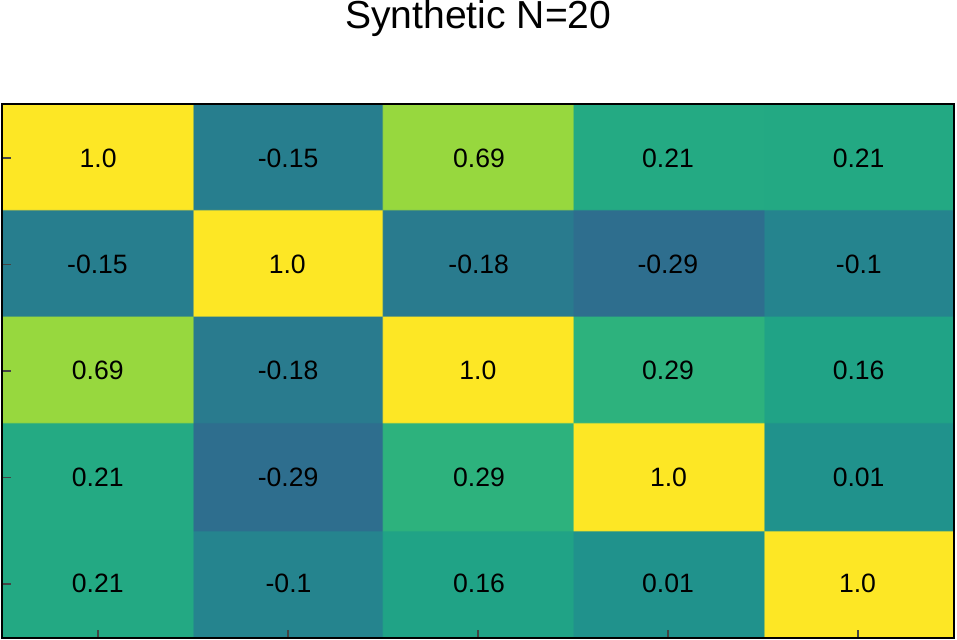}
}
\put(260,0)
{
\includegraphics[width=.33\textwidth, height=.33\textwidth]{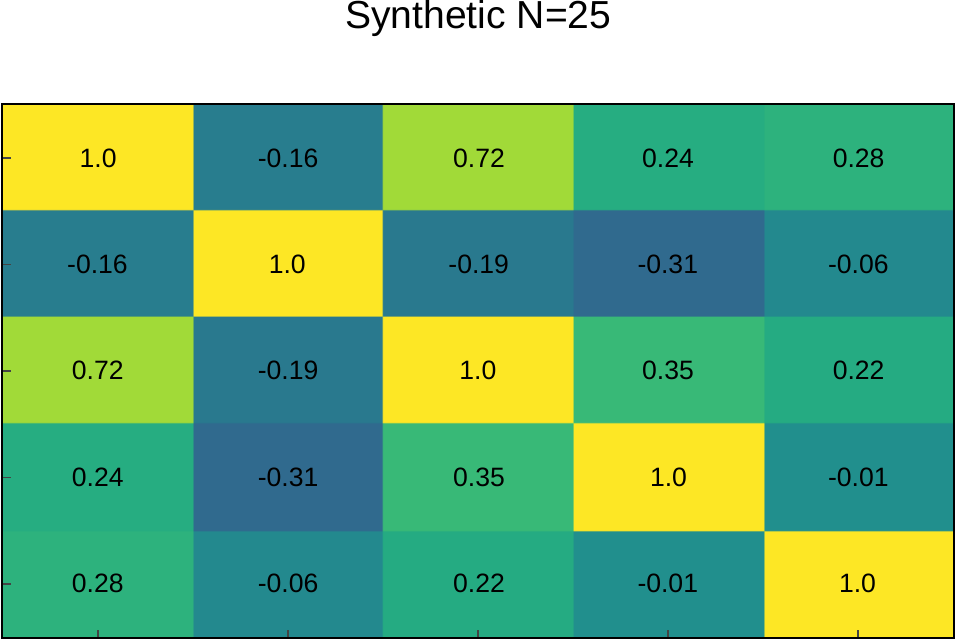}
}
\put(5,-25)
{
\includegraphics[width=\textwidth, height=0.05\textwidth]{colorbar.pdf}
}
\end{picture}
\caption{
Top panel: Pearson matrix of the original (\texttt{SustData}) dataset (left) and synthetic dataset (right) computed using first order conditional distribution with $N=25$. Bottom panel: Pearson matrix for synthetic data using first and second order conditional distributions with increasing $N$. The features in the bottom panel are ordered in the same manner as is displayed in the top panel. 
}
\label{fig:PearsonCorrOrder2}
\end{figure}

In Figure \ref{fig:PearsonCorrOrder1} and Figure \ref{fig:PearsonCorrOrder2} we present the Pearson correlation matrix for the original data and for various versions of the synthetic data. For the sake of clarity this is done only for five features of the datasets.
In Figure \ref{fig:PearsonCorrOrder1} we show the synthetic data generated using the first-order conditional probability as the granularity of the discretization is decreased. 
Figure \ref{fig:PearsonCorrOrder2} displays the corresponding plot utilizing the second-order conditional probability to generate the synthetic data.

%\begin{figure}[p] % Use [p] to place figures on a %dedicated page
%    \makebox[\hsize]{\includegraphics[width=1.4\textwidth, height=0.5\textheight, keepaspectratio]{PearCorrOrder1F5.png}}
%    \caption{Heat map of Pearson correlation matrix of the original dataset (top left) and the synthetic datasets computed with the first-order conditional probabilities at increasing resolution.}
%    \label{fig:PearsonCorrOrder1}% First figure
%    \vspace{0.5cm} % Optional spacing between images
%    \makebox[\hsize]{\includegraphics[width=1.4\textwidth, height=0.5\textheight, keepaspectratio]{PearCorrOrder2F5.png}}
%    \caption{Heat map of Pearson correlation matrix of the original dataset (top left) and the synthetic datasets computed with the first- and second-order conditional probabilities at increasing resolution.}
%    \label{fig:PearsonCorrOrder2} % Second figure
%\end{figure}

\begin{figure}
    \centering
    \includegraphics[width=12cm]{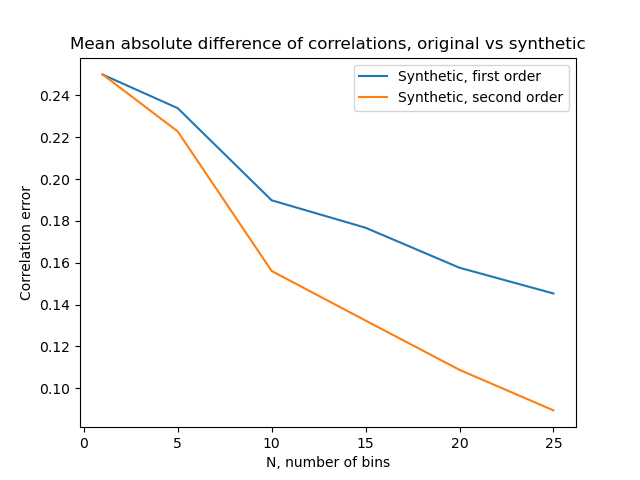}
    \caption{Plot of $\frac{1}{25}\sum_{i=1}^5 \sum_{i=j}^5 |C^\text{orig}_{ij} - C_{ij}|$, where $C^{\text{orig}}_{ij}$ and $C_{ij}$ are the entries in the respective Pearson correlation matrices as given in \eqref{eq:PearsonCorr}. This is the mean absolute difference of the Pearson correlation coefficients between original and synthetic datasets.}
    \label{fig:MeanPearCorr}
\end{figure}

We close this section by showing Figure \ref{fig:MeanPearCorr}. 
In Figure \ref{fig:MeanPearCorr} the mean absolute error between the Pearson correlation coefficients of the original dataset and the synthetic dataset are shown for both first and second-order conditional probabilities as a function of $N$. 
Figure \ref{fig:condDistEx750}, Figure \ref{fig:condDistOrder2}, and Figure \ref{fig:MeanPearCorr} taken together indicate that the method using the second-order conditional probability distributions does a better job in preserving the correlations, 
even with this simplistic choice of discretization of the features (uniform discretization, small value of $N$).
The behaviors of the two different synthetic datasets in Figure \ref{fig:MeanPearCorr} are very similar as $N$ increases. 
If this trend persists as one increase the depth of conditional probabilities, it suggests that the two parameters $N$ and the depth of conditional probabilities can indeed be used to tune how well the synthetic dataset should represent the original, assuming that the green and orange curves in Figure \ref{fig:MeanPearCorr} will not cross as $N$ increases further.

\section{Concluding remarks}\label{final_discussion}
\NJ{In this paper we use simple statistics to construct a algorithmic procedure capable to produce synthetic data with tunable distributional and correlational fidelity for tabular data. The purpose for this construction is to try and meet the competing requirements of privacy and statistical relevance (of data owners and data users).}

\NJ{Motivated by the need of researchers to obtain energy-related data for forecasting purposes, we look at a specific large original dataset, taken from \cite{pereira_sustdata}, see also \cite{Pereira2014}.} We create many synthetic versions of it, having the same size and features with different entries. 
We preserved to some extent in the synthetic dataset $\mathcal{S}$ the correlations between features of the original dataset $\mathcal{O}$.

In future work we plan investigate the connection between this gap and the level of privacy offered by this method. A key message to take away is that our investigation shows that the two parameters $N$ and the depth of conditional probabilities can be used to tune how well the synthetic dataset should represent or hide the original dataset. From Figure \ref{fig:PearsonCorrOrder1} and Figure \ref{fig:PearsonCorrOrder2} it is clear that within our current investigation, the range of tunability using $N$ depends on the order of conditional probabilities used. \NJ{From the observed results it is reasonable to expect that utilizing deeper conditional distributions would enlarge the space of tunability when generating synthetic data. This tunability has the potential to facilitate data sharing between data owners and users, relying on algorithms that are understandable and transparent.}

Quite interestingly, a few innovative ideas for quantifying rigorously privacy (either differential, metric, or something else) in terms of error bounds exist; see e.g. \cite{Snoke2018,Dankar2022,Yuan2023,Boedihardjo} and references cited therein. 
We plan to explore in  the near future to which extent such ideas are applicable to our context.
\clearpage
\subsection*{Funding}
N.J, J.F, and A.M. are involved in Swedish Energy Agency's project Solar Electricity Research Centre (SOLVE) with grant number 52693-1. R.L. and A.M. are grateful to Carl Tryggers Stiftelse for their financial support via the grant CTS 21:1656. A.M. thanks NAISS for the project 2023/22-1283, which provided the needed  computational resources.

\subsection*{Authors' contributions}
\textbf{NJ}: Methodology, Software, Data Curation, Writing - Original Draft, Visualization.
\textbf{AM}: Conceptualization, Supervision, Writing - Review \& Editing.
\textbf{RL}: Methodology, Resources, Writing - Review \& Editing.
\textbf{JF}: Conceptualization, Writing - Review \& Editing. 

\subsection*{Competing Interests}
We have no competing interests, out results and implementations are open.

\subsection*{Acknowledgments}
We thank Grigor Nika (Karlstad, Sweden) for fruitful initial discussions on the synthetic data topic. We also would like to thank Marieke Rynoson (Falun, Sweden) for helpful notes on the manuscript.

%\newpage
\printbibliography

\end{document}